\documentclass[12pt]{article}
\usepackage{setspace}
\usepackage[round]{natbib}
\usepackage{tcolorbox}              
\usepackage[font=small]{caption}
\usepackage{subcaption}
\usepackage{setspace}
\usepackage{pdfpages}
\usepackage{hyperref}
\hypersetup{colorlinks,
linkcolor={blue},
citecolor={blue},
urlcolor={red}}
\usepackage[utf8]{inputenc}
\usepackage[toc,page]{appendix}

\title{\textbf{Finding patterns in Knowledge Attribution for Transformers}}
\author{
\textcolor{red}{Jeevesh Juneja}\\
Delhi Technological University\\
\texttt{creativityinczenyoga@gmail.com}\\
\and
\textcolor{red}{Ritu Agarwal}\\
Delhi Technological University\\
\texttt{rituagarwal@dtu.ac.in}}

\begin{document}

\maketitle

\begin{abstract}
    We analyze the Knowledge Neurons\citep{dai2021knowledge} framework for the attribution of factual and relational knowledge to particular neurons in the transformer network. We use a 12-layer multi-lingual BERT model for our experiments. Our study reveals various interesting phenomena. We observe that mostly factual knowledge can be attributed to middle and higher layers of the network($\ge 6$). Further analysis reveals that the middle layers($6-9$) are mostly responsible for relational information, which is further refined into actual factual knowledge or the "correct answer" in the last few layers($10-12$). Our experiments also show that the model handles prompts in different languages, but representing the same fact, similarly, providing further evidence for effectiveness of multi-lingual pre-training. Applying the attribution scheme for grammatical knowledge, we find that grammatical knowledge is far more dispersed among the neurons than factual knowledge. 
    
\end{abstract}
\section{Introduction}
\pagenumbering{arabic}

\cite{geva_kvm} showed that the feed-forward layers of the transformer \citep{vaswani} which constitute almost two-thirds of the model's parameters operate as a key-value store, with each key correlating with textual patterns which the model has successfully learned to detect. The values corresponding to these keys are indicative of the output distributions over vocabulary of the model for that particular pattern. 

\cite{dai2021knowledge} extends this to relate "facts" (for e.g. that capital of France is Paris) to certain key-value memories, i.e., to certain neurons in the feed-forward layers of the transformer model. They conduct experiments which show that modulating these particular neurons' activations has much larger impacts on the correct output's probabilities, as compared to the case where they modulate random neurons' activations. 

In the reverse direction, they also show that queries related to the fact tend to activate these particular neurons more than others. Finally, they provide methods to update and erase preexisting knowledge/facts in these neurons without any fine-tuning.

We explore their attribution strategy further, with different kinds of knowledge, at different layers of the network, proposing hypotheses for inner working of transformer language models. 

\section{Related Works}
The concept of "memory" of a model has been one that has existed for a long time and has helped push the state-of-the-art of language models further consistently(\cite{hopfield}, \cite{lstm}). As the models become larger and larger, they store more and more memories, but it becomes harder to learn and retrieve them reliably. 

Thus, much recent work has focused on decoupling memory storage from model learning and on new ways to access memories. \cite{mem_nets}, \cite{e2e_mem_nets}, try to augment the memory of recurrent models, while \cite{lample2019large}, \cite{rac_nntke} are recent attempts to integrate and extend the memory of transformer networks efficiently. 

We believe understanding the existing knowledge storage and retrieval mechanisms of transformers, will pave the way for better and efficient integration of external memory stores, and models that can store more memories, more efficiently. 
Recent works in machine learning interpretability(\cite{voss2021visualizing}, \cite{olah2018the}) try to describe the attribution of various features and predictions to various layers and weights for various computer vision models such as ResNet\citep{resnet}. To the best of our knowledge, this is the first work to explore the attribution distributions across different layers and types of knowledge in natural language processing.

\section{Types of Knowledge}
This paper deals with three types of knowledge, which are explained with the following example.
\begin{tcolorbox}
"Sarah was visiting [MASK1], the capital [MASK2] France"
\end{tcolorbox}

\textbf{Factual knowledge} refers to knowledge regarding a certain fact. In the above example, it refers to whether the model can predict the correct word "Paris" at masked position 1, with high probability or not.

\textbf{Relational Knowledge} refers to the knowledge regarding a relation. In the above example, it refers to whether a model is able to identify that it is being asked for the "capital" of \textit{some} nation.

\textbf{Grammatical Knowledge} refers to whether a model knows about the grammatical structure of the language on which it was trained. In the above example it refers to whether the model can predict the correct proposition("of") at masked position 2, with high probability.

This paper concerns itself with the following three questions:

\begin{enumerate}
    \item How exactly is attribution for Factual knowledge distinct from the attribution of relational knowledge? Which layers are responsible for which knowledge?\\
    \item Do facts in different languages correspond to the same neurons? That is, whether a language model trained on multi-lingual data, has a common representation space for those languages?\\
    \item How is grammatical knowledge dispersed in the various layers of the transformer model? How is this different from relational or factual knowledge?
\end{enumerate}

\section{Attribution Strategy for Facts}
We prompt the model, using masked sentences, similar in style to the one shown before. Following which, we inspect the logit corresponding to the correct prediction, (i.e., Paris for [MASK1]), and attribute the logit's value to the various neurons in the transformer network using the integrated gradients strategy \citep{integrated_grads} as used in \cite{dai2021knowledge}. 

This method provides an attribution scheme that is sensitive to changes in input, but invariant to changes in the implementation of the model. At its core, it involves selecting a baseline input($x'$) and then integrating the gradients of the output($F(x'+\alpha(x-x')$), with respect to the input, as we increase the input from the baseline(at $\alpha=0$) to its actual value(at $\alpha=1$). In our case, the baseline input is chosen as 0, for every neuron.

\begin{equation}
    IntegratedGrads_i(x)::= (x_i-x_i') \times \int_{\alpha=0}^1 \frac{\partial F(x'+\alpha(x-x'))}{\partial x_i} d\alpha
    \label{eqn:integrated_grads}
\end{equation}

To find the contribution of a single neuron($i$) to the final prediction of, say Paris, we integrate the gradients of the output logit corresponding to Paris, with respect to the activation of that neuron. As we can't compute the integration in Equation~\ref{eqn:integrated_grads} exactly, we compute a discrete sum approximation of the integration. After finding the attribution scores of each neuron toward the final prediction this way, we find neurons with maximum or near-maximum scores and the results of \cite{dai2021knowledge} show that these are the ones that store the information leading to the correct prediction.

Furthermore, to remove false positives and identify neurons corresponding to actual facts rather than some spurious signal in the input, \cite{dai2021knowledge} suggest using multiple prompts(corresponding to the same fact) instead of just a single one and only choosing those neurons which have high attribution scores for most of the prompts. These prompts need to have varied structures and grammatical syntax. For example, another prompt corresponding to the (France, Capital, Paris) fact can be:

\begin{tcolorbox}
France's capital, [MASK] is a hotspot for romantic vacations.
\end{tcolorbox}

We use the implementation\footnote{\url{https://github.com/EleutherAI/knowledge-neurons}} released by EleutherAI, to run our experiments.

\section{Initial Exploration}
\begin{figure}
    \centering
    \includegraphics[scale=0.75]{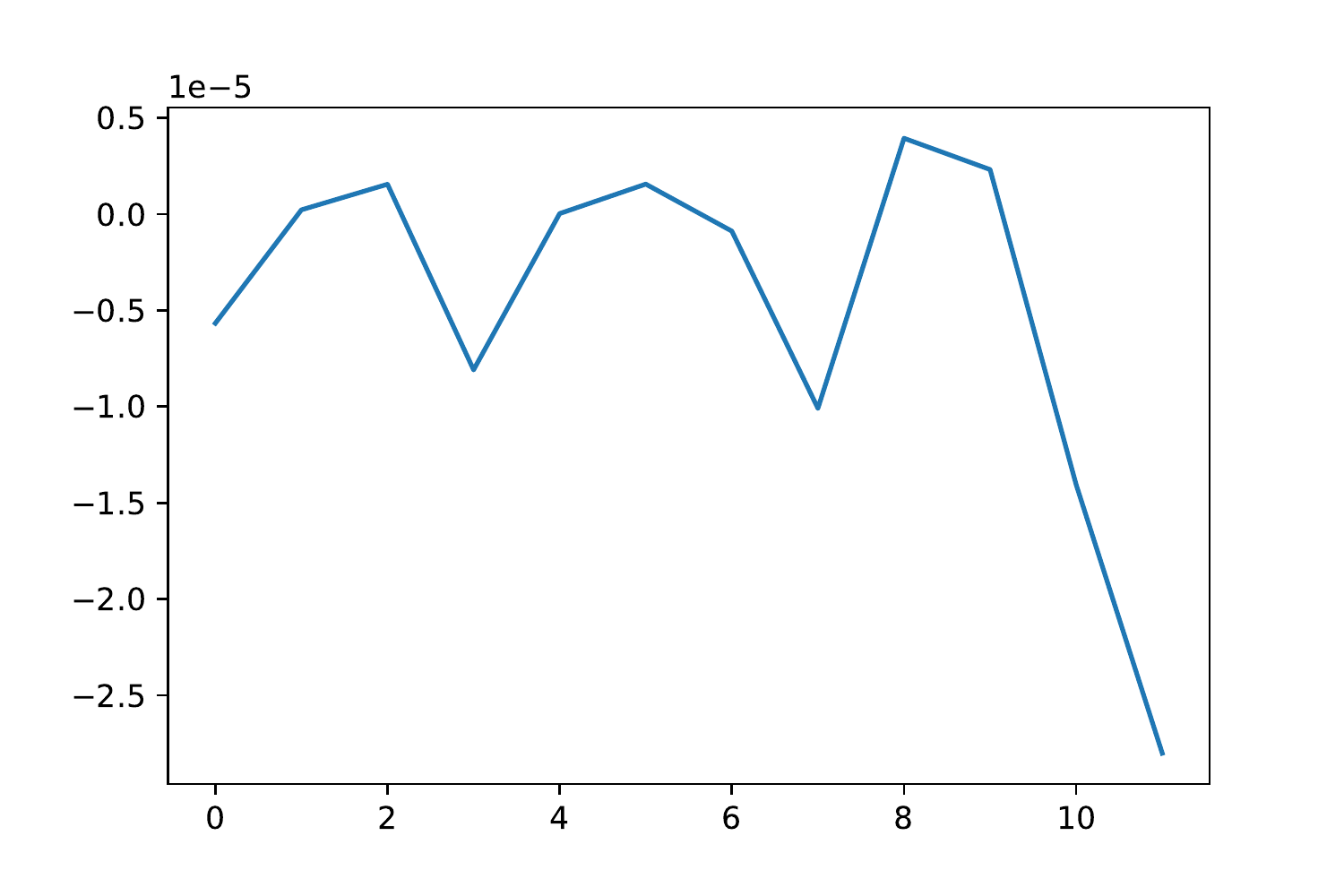}
    \caption{Mean of layer-wise average scores over prompts corresponding to (France, Capital, Paris). Shows the mean attribution of the fact that Paris is capital of France to various layers of the transformer model. The layer number varies across X=axis and mean score along Y-axis.}
    \label{fig:avg_scores_fr}
\end{figure}

Figure~\ref{fig:avg_scores_fr} shows the average scores over the multiple prompts\footnote{See Appendix~\ref{appendix_first} for a complete list of prompts.} considered, for each layer of the model. We see that most initial layers consist of approximately 0 attribution scores. Layers 7-8-9 yield high attribution scores, followed by huge(in comparison) negative scores for the final layers. We hypothesize that this behavior is due, mainly to three reasons:

\begin{itemize}
    \item Initial layers encode low-level syntactic information, which doesn't particularly correspond to the fact of interest.
    \item The layers 7-8-9 encode higher level information that is of interest to prediction for the fact of our interest, possibly of the relation(viz., "capital") that we are seeking.
    \item The last few layers, specialize in lots of different directions and only few neurons correspond to the actual fact of interest. The rest of them try to fit the lower layer representations to the fact they contain(remember, each neuron is a key-value memory) and hence lead to lower attribution score for the actual fact of interest.
\end{itemize}

\begin{figure}
    \centering
    \includegraphics[scale=0.75]{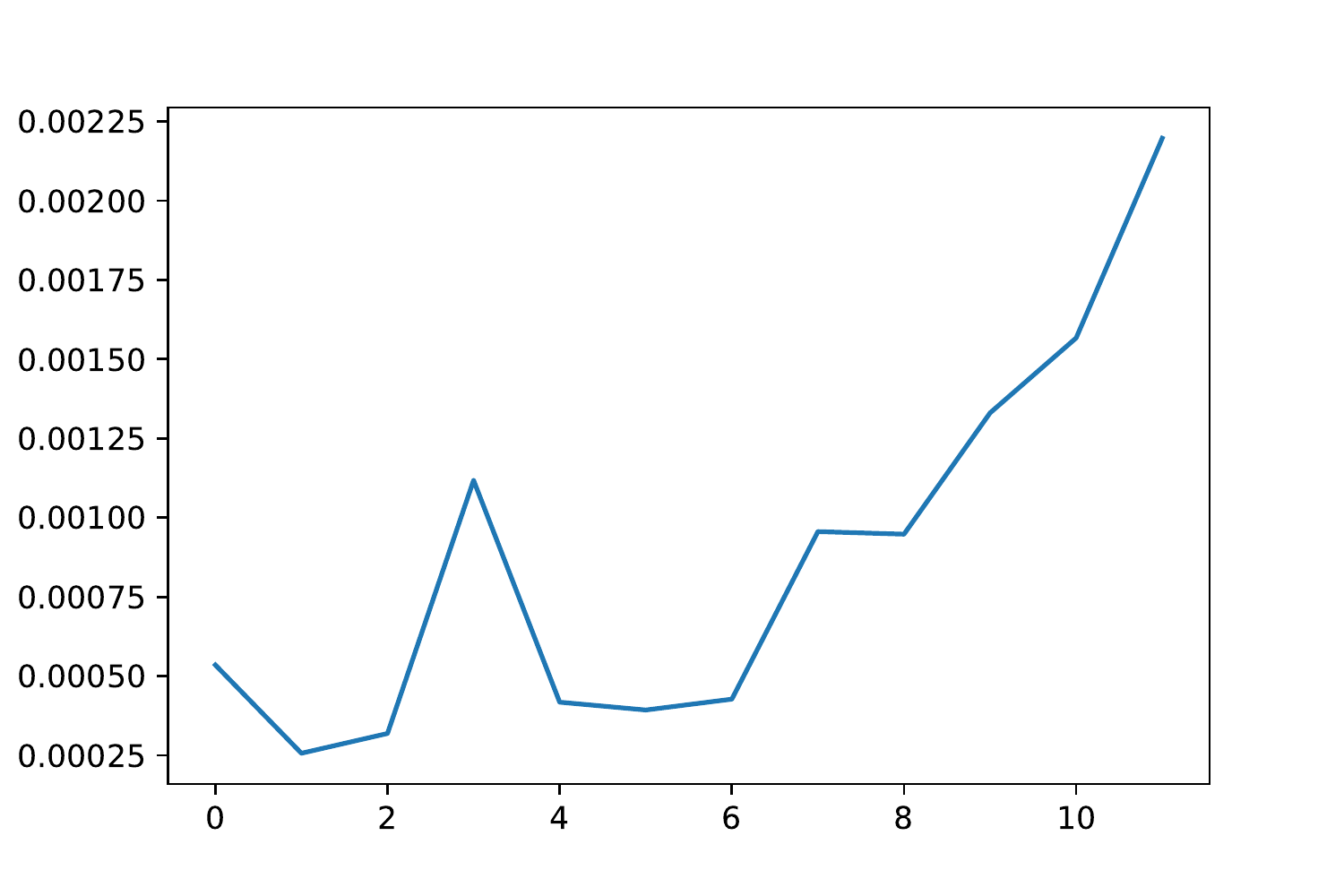}
    \caption{Mean of layer-wise standard deviations of scores over prompts corresponding to (France, Capital, Paris). Layer number on X-axis, standard deviation on Y-axis.}
    \label{fig:avg_std_scores_fr}
\end{figure}

The sharp increase in standard deviations from the mean, in the last few layers, in Figure~\ref{fig:avg_std_scores_fr} further supports our claim that the last few layers store a variety of different facts. 

\begin{figure}
    \centering
    \includegraphics[scale=0.75]{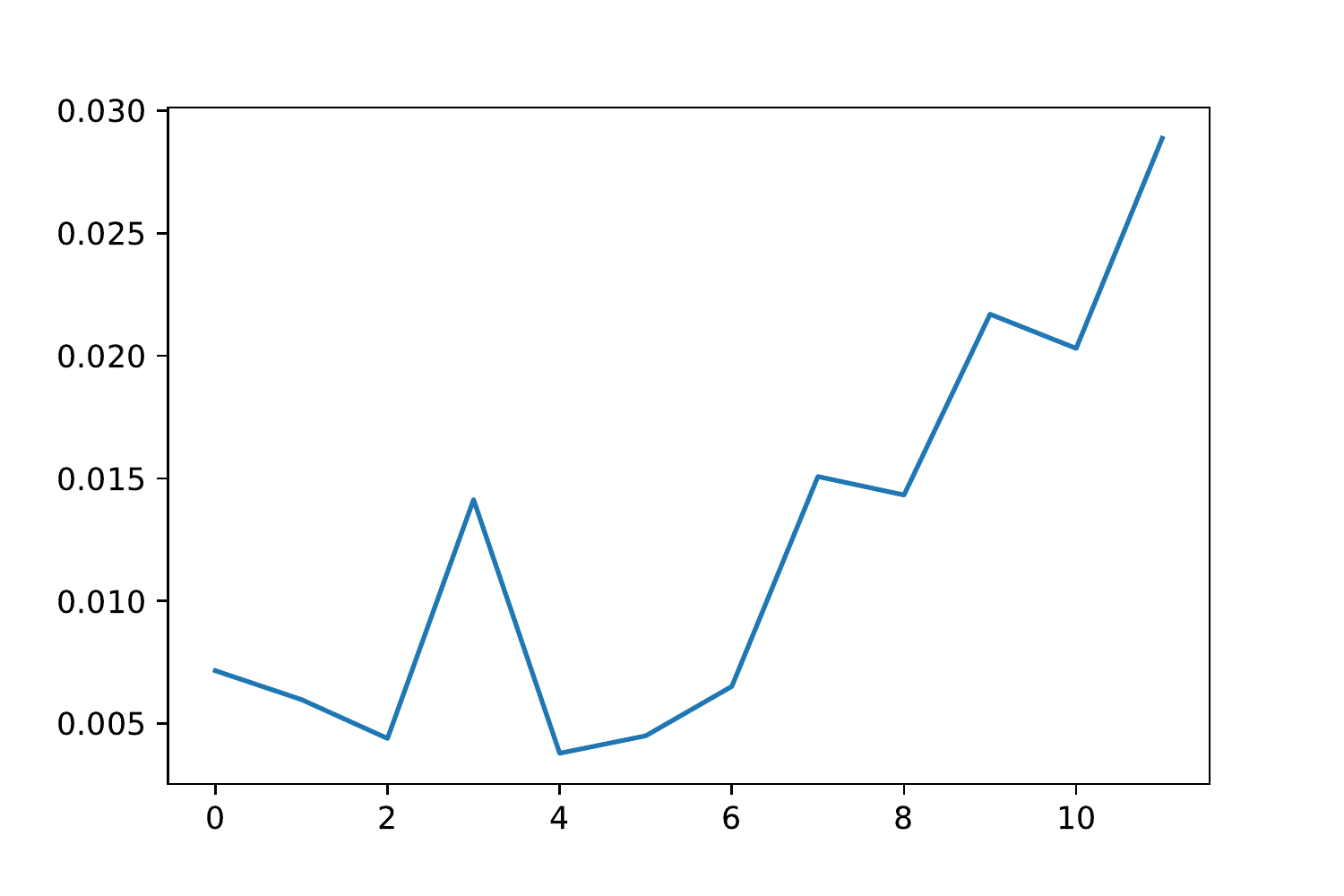}
    \caption{Mean of layer-wise maximum of scores over prompts corresponding to (France, Capital, Paris).Layer number on X-axis, maximum scores on Y-axis.}
    \label{fig:max_scores_fr}
\end{figure}

Figure~\ref{fig:max_scores_fr} provides further insight into how knowledge is distributed and retrieved from the transformer. We see that although the mean score attains its maximum around layer 8, the maximum score goes on increasing as we go higher in the model. This suggests that although most of the neurons in higher layers store information highly unrelated to our fact, there does exist some neuron(the one with maximum attribution) that stores exactly the information we need to predict the correct value.

\begin{figure}
    \centering
     \begin{subfigure}[b]{0.3\textwidth}
         \centering
         \includegraphics[width=\textwidth]{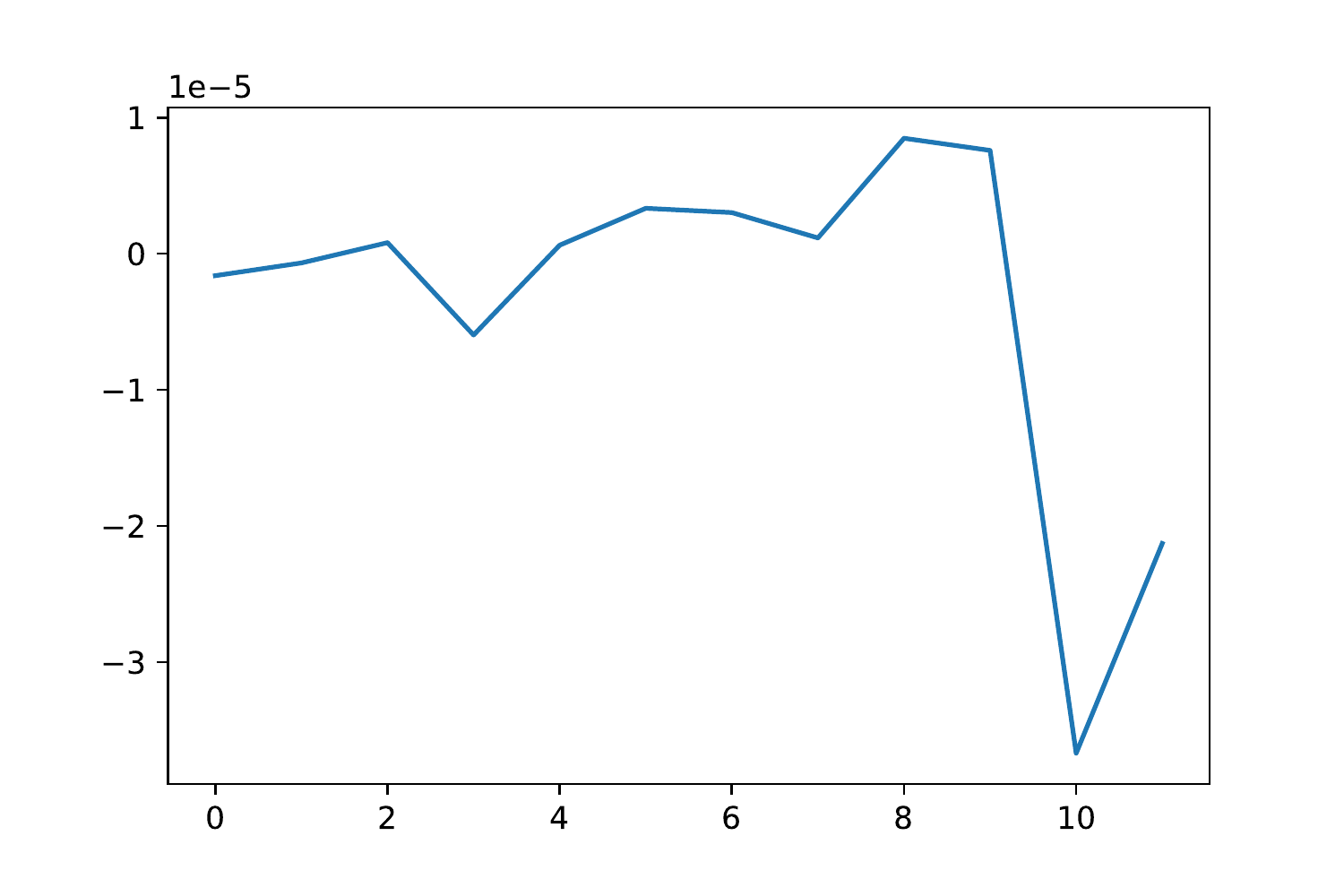}
         \caption{Average scores}
         \label{fig:mean_scores_de}
     \end{subfigure}
     \hfill
     \begin{subfigure}[b]{0.3\textwidth}
         \centering
         \includegraphics[width=\textwidth]{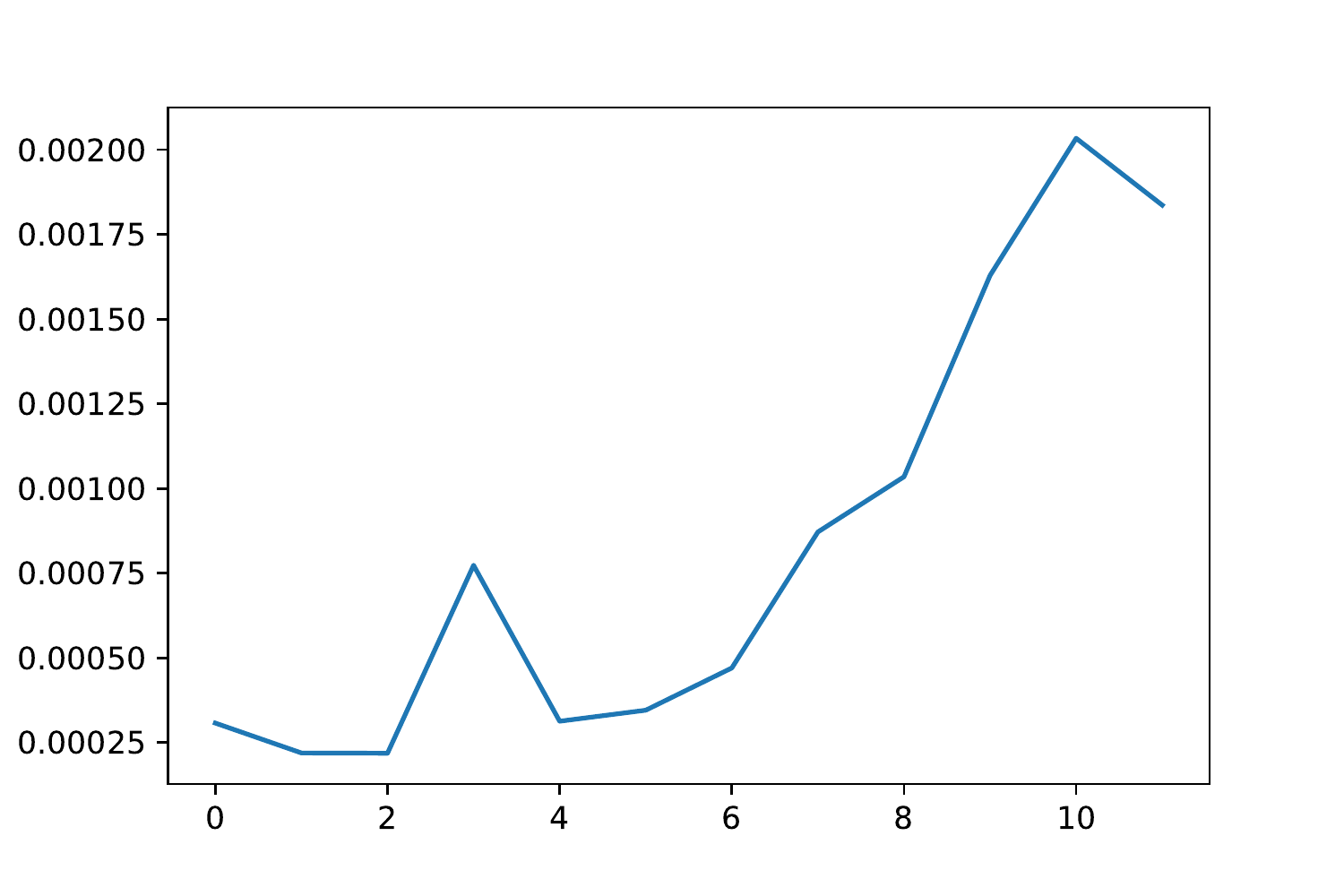}
         \caption{Standard deviations}
         \label{fig:std_scores_de}
     \end{subfigure}
     \hfill
     \begin{subfigure}[b]{0.3\textwidth}
         \centering
         \includegraphics[width=\textwidth]{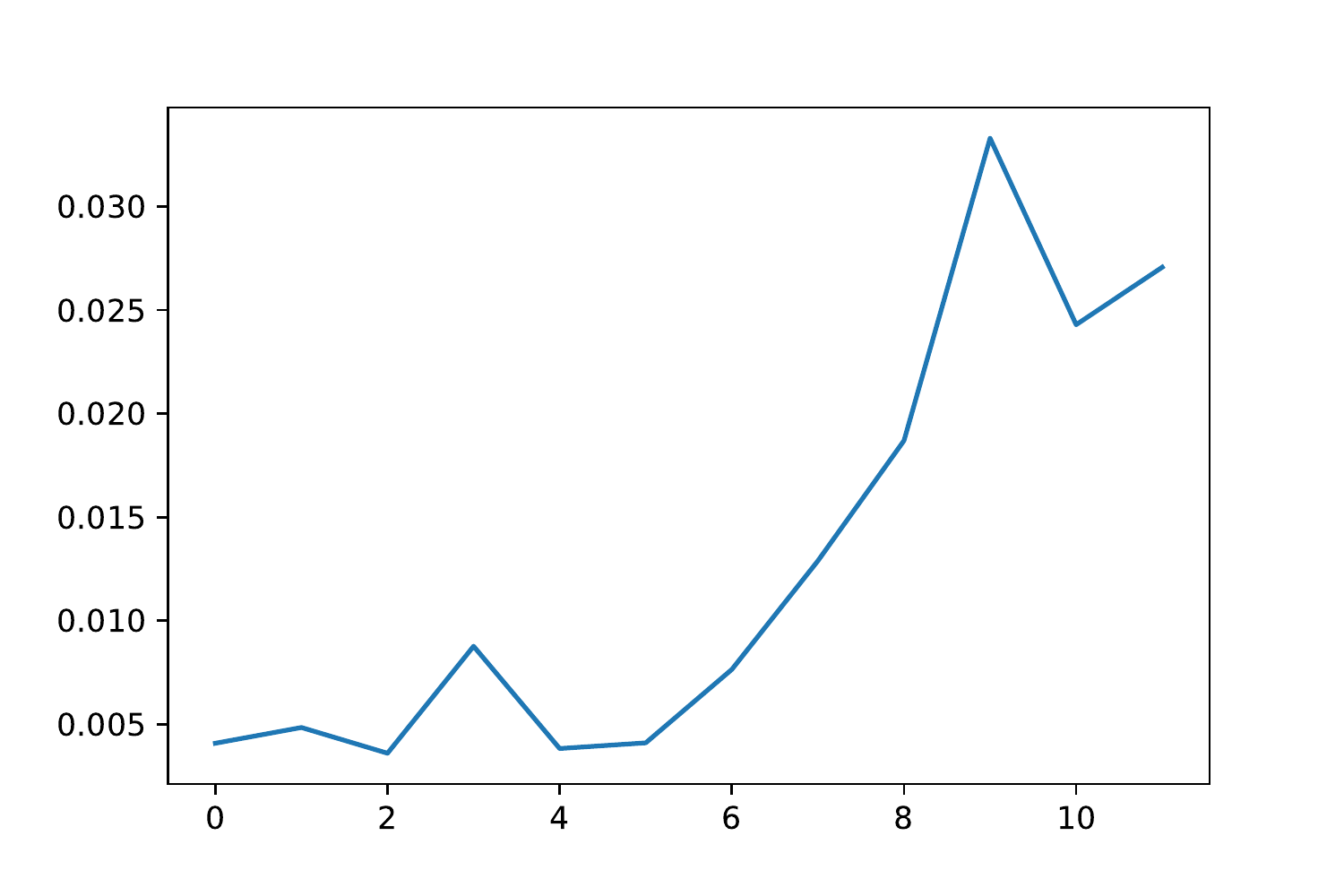}
         \caption{Maximum scores}
         \label{fig:max_scores_de}
     \end{subfigure}
        \caption{Variations of information content(attribution scores) at various layers for prompts corresponding to the fact (Germany, Capital, Berlin).}
        \label{fig:de_scores}
\end{figure}

In Figure~\ref{fig:de_scores} we see the same kind of behavior for prompts that correspond to a different fact, showing the generalizability of our claims. 

\section{Is it the relation or the fact?}

Next, we explore whether neurons we find using this attribution method correspond to the fact that "Paris is the capital of France" or the relation between the two objects, i.e., "A is capital of B". \cite{dai2021knowledge} use a threshold $t$, and only neurons with attribution scores$> t$ are considered the ones corresponding to this fact. In addition, they use another threshold $P$, and only neurons that have $> t$ attribution scores in more than $P\%$ of the prompts are selected as representing the fact.  

\begin{figure}
    \centering
    \includegraphics[scale=0.75]{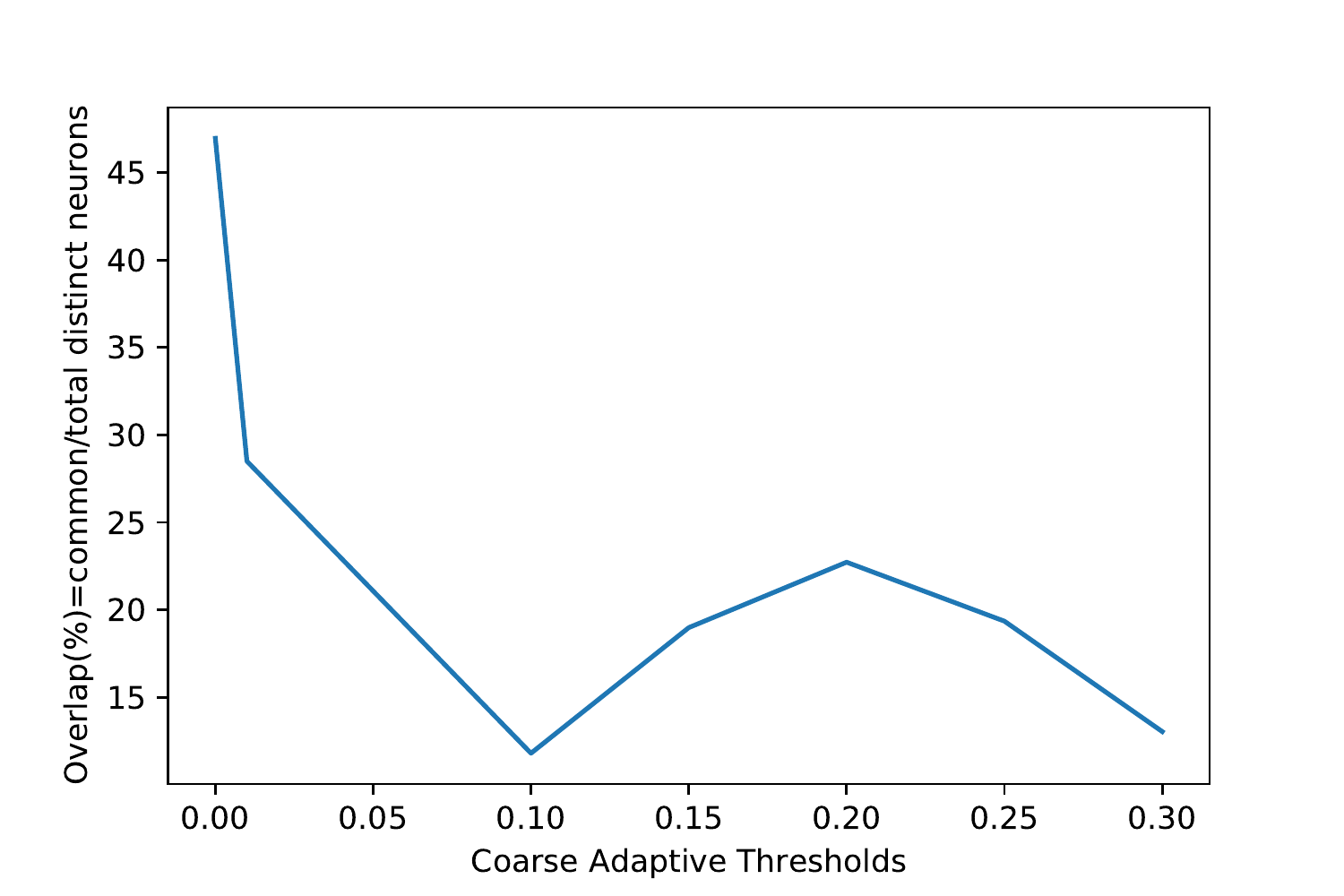}
    \caption{Overlap between knowledge neurons for (France, Capital, Paris) \& (Germany, Capital, Berlin) with varying $t$.}
    \label{fig:relation_or_fact_fr_de}
\end{figure}

We fix $P$ to $50\%$ and vary $t$ to find how the overlap between the neurons identified as representing the fact (France, Capital, Paris) and (Germany, Capital, Berlin) changes. Increasing $t$ for a particular fact has the following effect: at $t=0$ we are selecting all neurons of the network, with increasing $t$ we start selecting neurons that are more and more particular to the fact. If we continue to increase $t$ further, it is likely that we will lose some of the meaningful neurons as well. 

The results are shown in Figure~\ref{fig:relation_or_fact_fr_de}. At small thresholds the overlap is quite high(as expected) because we are selecting a large number of neurons for each fact(even unrelated neurons get selected), but the bump in the percentage of common neurons, {\bf with increased threshold, $t$} clearly indicates that there are neurons storing the commonality, that is, "A is capital of B", rather than individual information regarding Paris being capital of France or Berlin being capital of Germany. 

\begin{figure}
    \centering
    \includegraphics[scale=0.75]{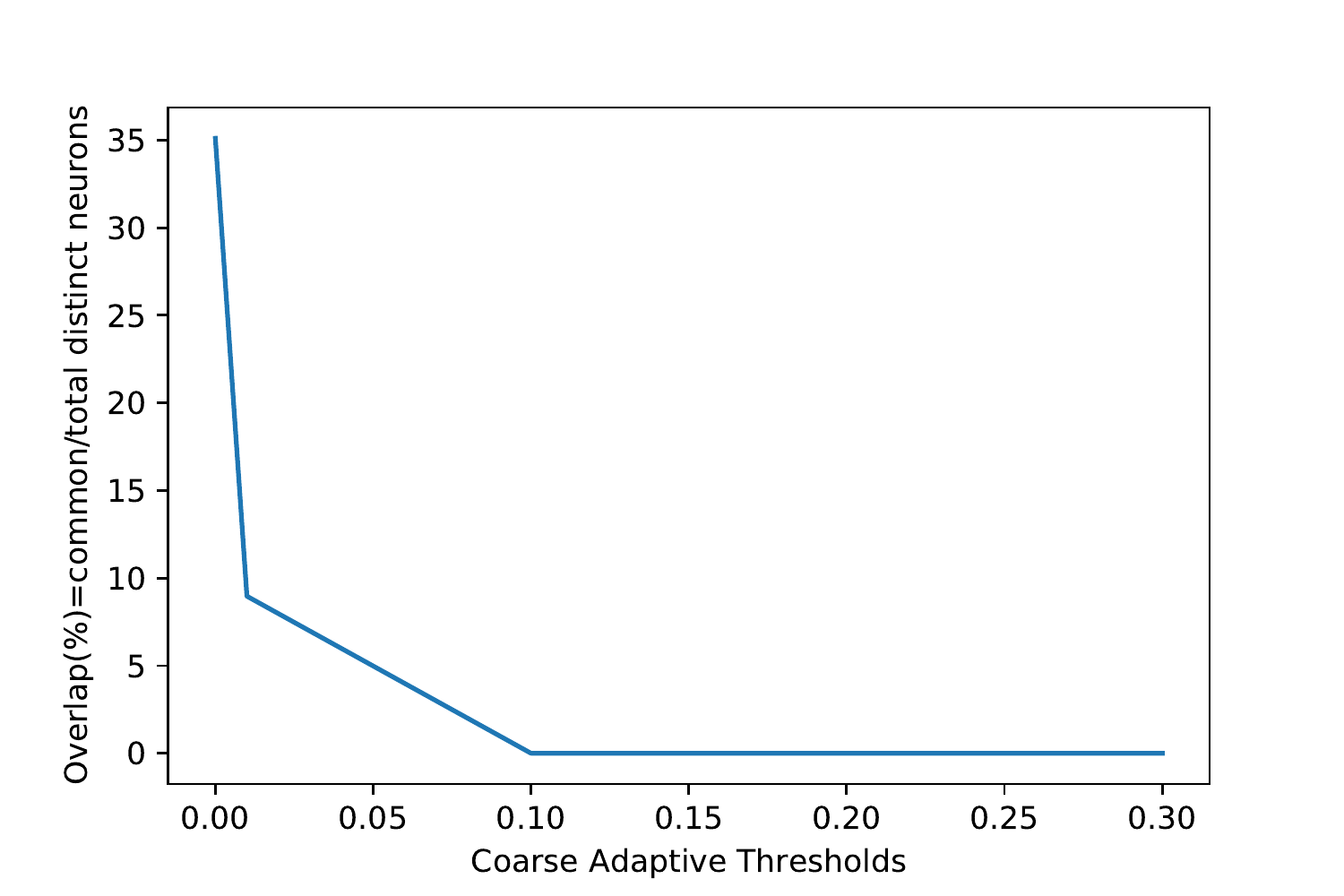}
    \caption{Overlap between knowledge neurons for (France, Capital, Paris) \& (Cow, eats, grass) with varying $t$.}
    \label{fig:relation_or_fact_fr_cow}
\end{figure}

Moreover, predictions at masked positions, for both of these kinds of prompts, are being based on these common "relational" neuron. For comparison, we provide the overlap of commons neurons for two unrelated facts, viz. (France, Capital, Paris) and (Cow, eats, grass), in Figure~\ref{fig:relation_or_fact_fr_cow}. We see no bump and instead observe a flat line here.

\begin{figure}
    \centering
     \begin{subfigure}[b]{0.45\textwidth}
         \centering
         \includegraphics[width=\textwidth]{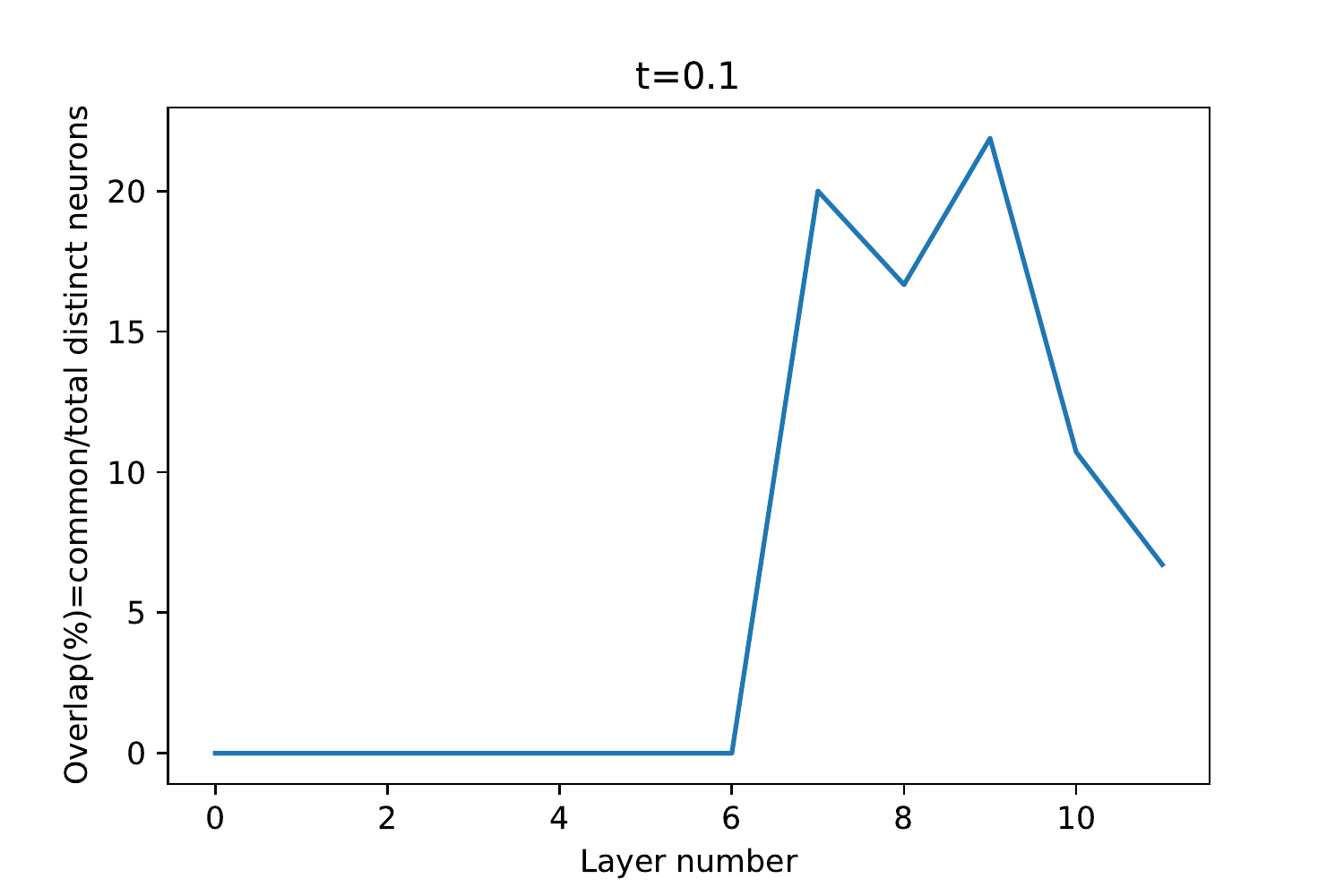}
         \caption{$t=0.1$}
         \label{fig:overlap_1}
     \end{subfigure}
     \hfill
     \begin{subfigure}[b]{0.45\textwidth}
         \centering
         \includegraphics[width=\textwidth]{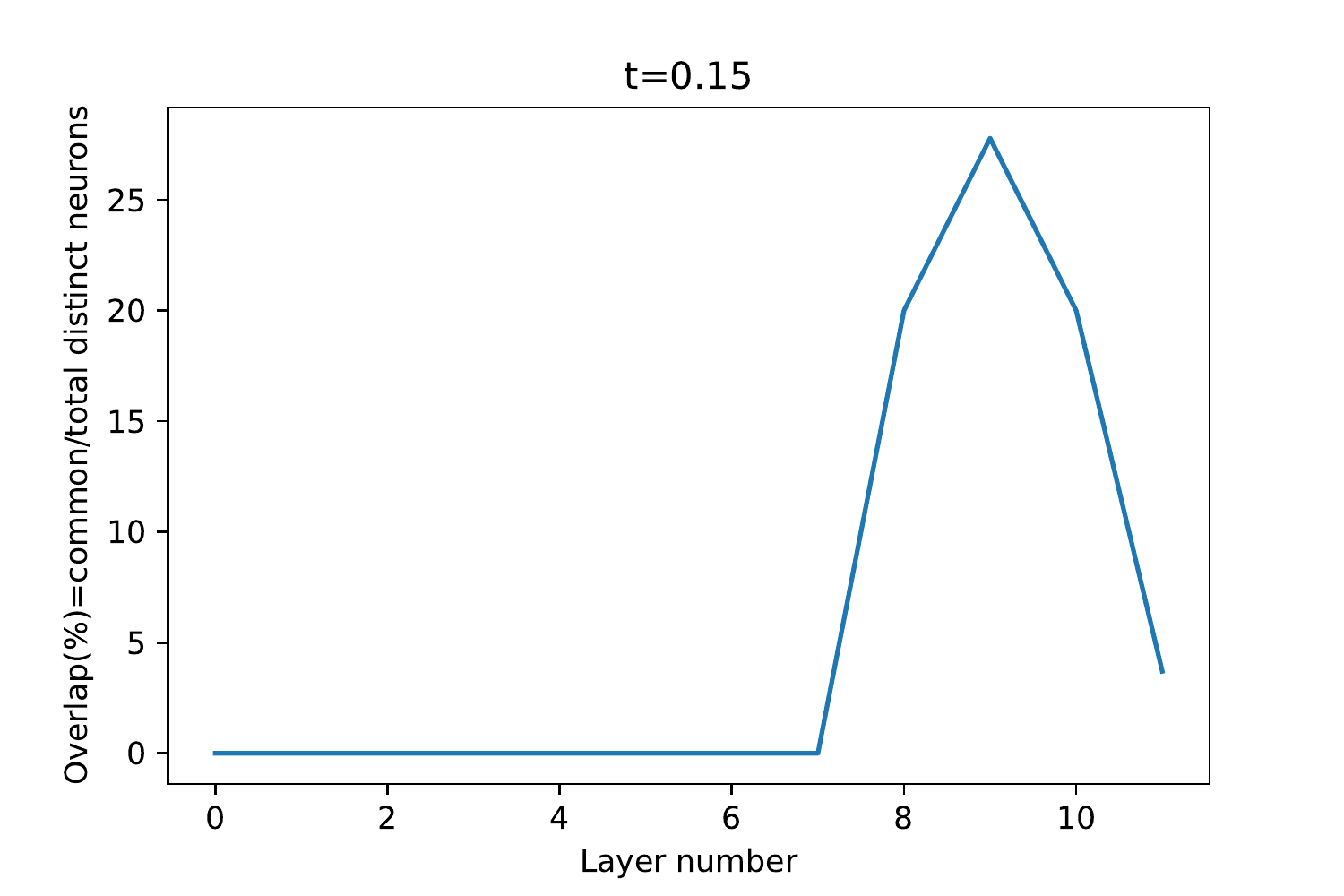}
         \caption{$t=0.15$}
         \label{fig:overlap_2}
     \end{subfigure}
     \hfill
     \begin{subfigure}[b]{0.45\textwidth}
         \centering
         \includegraphics[width=\textwidth]{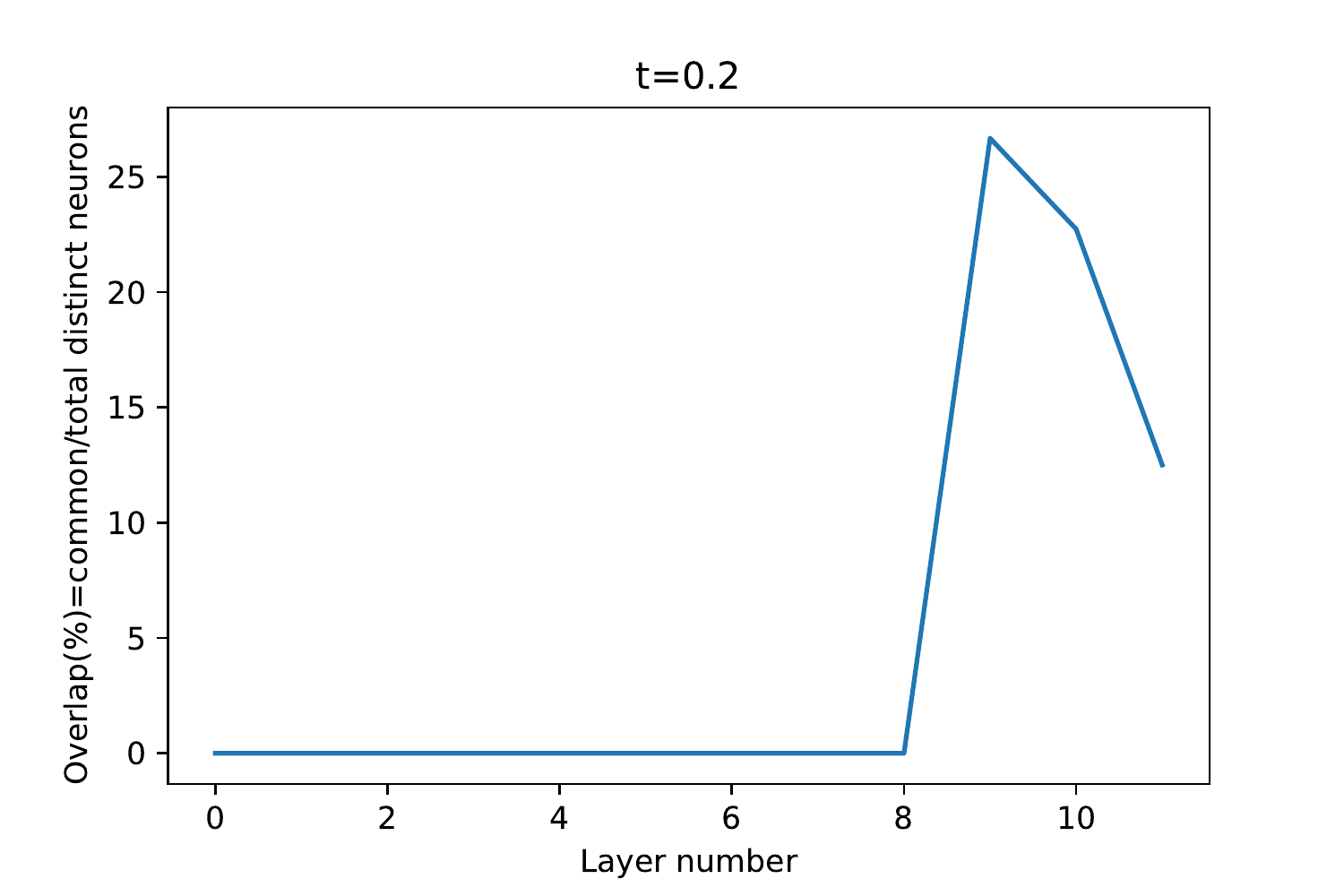}
         \caption{$t=0.2$}
         \label{fig:overlap_3}
     \end{subfigure}
     \hfill
     \begin{subfigure}[b]{0.45\textwidth}
         \centering
         \includegraphics[width=\textwidth]{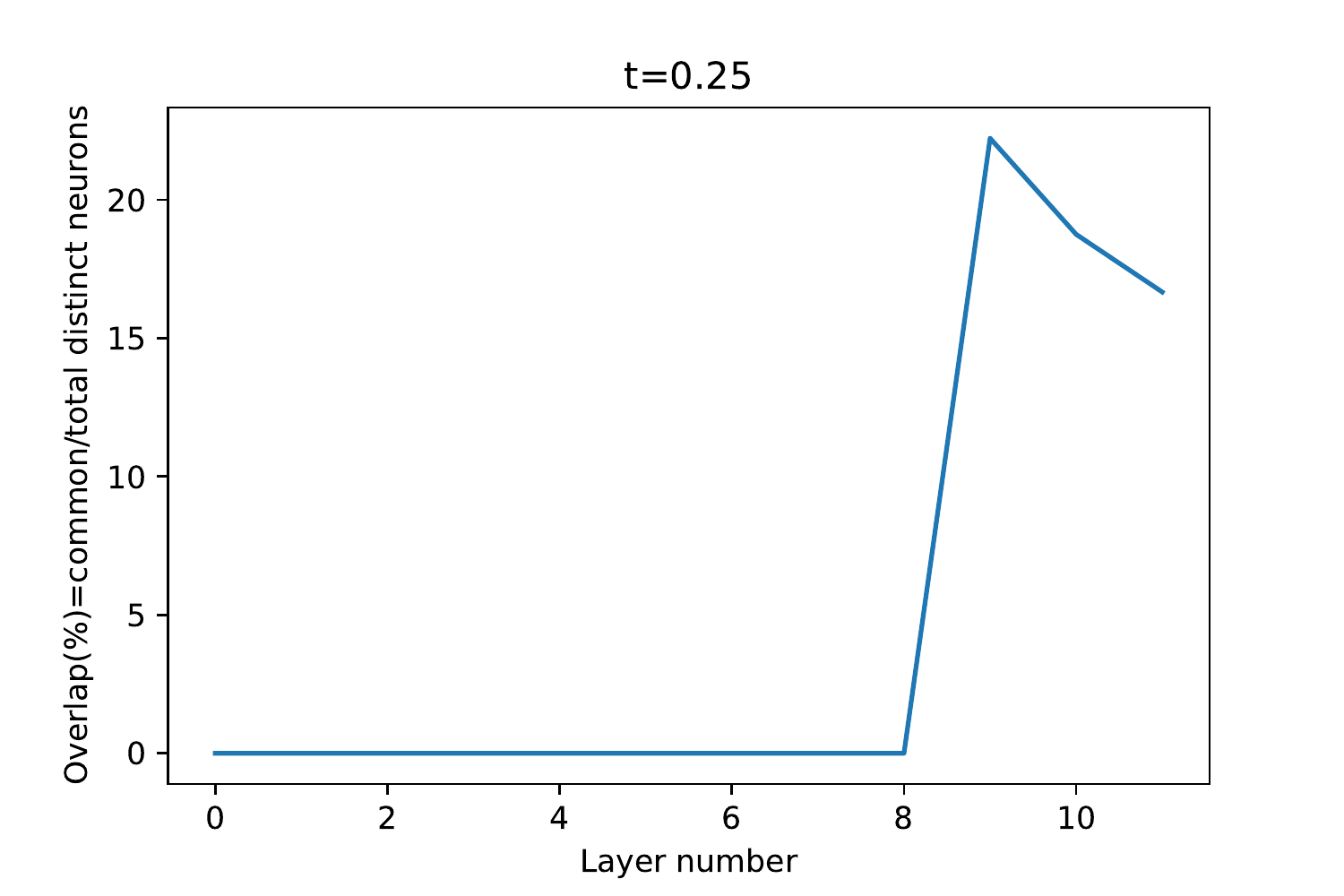}
         \caption{$t=0.25$}
         \label{fig:overlap_4}
     \end{subfigure}
        \caption{Layer-wise overlap of neurons between the facts (Germany, Capital, Berlin), and (France, Capital, Paris) at different $t$ in the bump in Figure~\ref{fig:relation_or_fact_fr_de}.}
        \label{fig:overlap_at_layers_with_t}
\end{figure}

In Figure~\ref{fig:overlap_at_layers_with_t}, we see exactly in which layers the common neurons lie. Overall, we observe a trend that:
\begin{itemize}
    \item No model picks any neurons in lower layers
    \item Most common neurons are found in layer 7-8-9-10. It is likely that the "capital" relation is being realised by the model in these layers. 
    \item In higher layers, we observe a drop in the number of common neurons, again indicating that each fact must belong to different set of neurons in higher layer. It is likely that the model is finding the exact capital of the specified country.
\end{itemize}
\section{Do same neurons represent the same fact in different languages?}
We use a multilingual transformer encoder model, {\bf m-BERT} \citep{m-bert} to check whether facts in different languages access information from the same neurons? We construct two sets of prompts corresponding to the relational-fact (France, Capital, Paris) in English, and translate one set of prompts to French to construct a set of prompts corresponding to the same fact, but in a different language.

\begin{figure}
    \centering
    \includegraphics[scale=0.75]{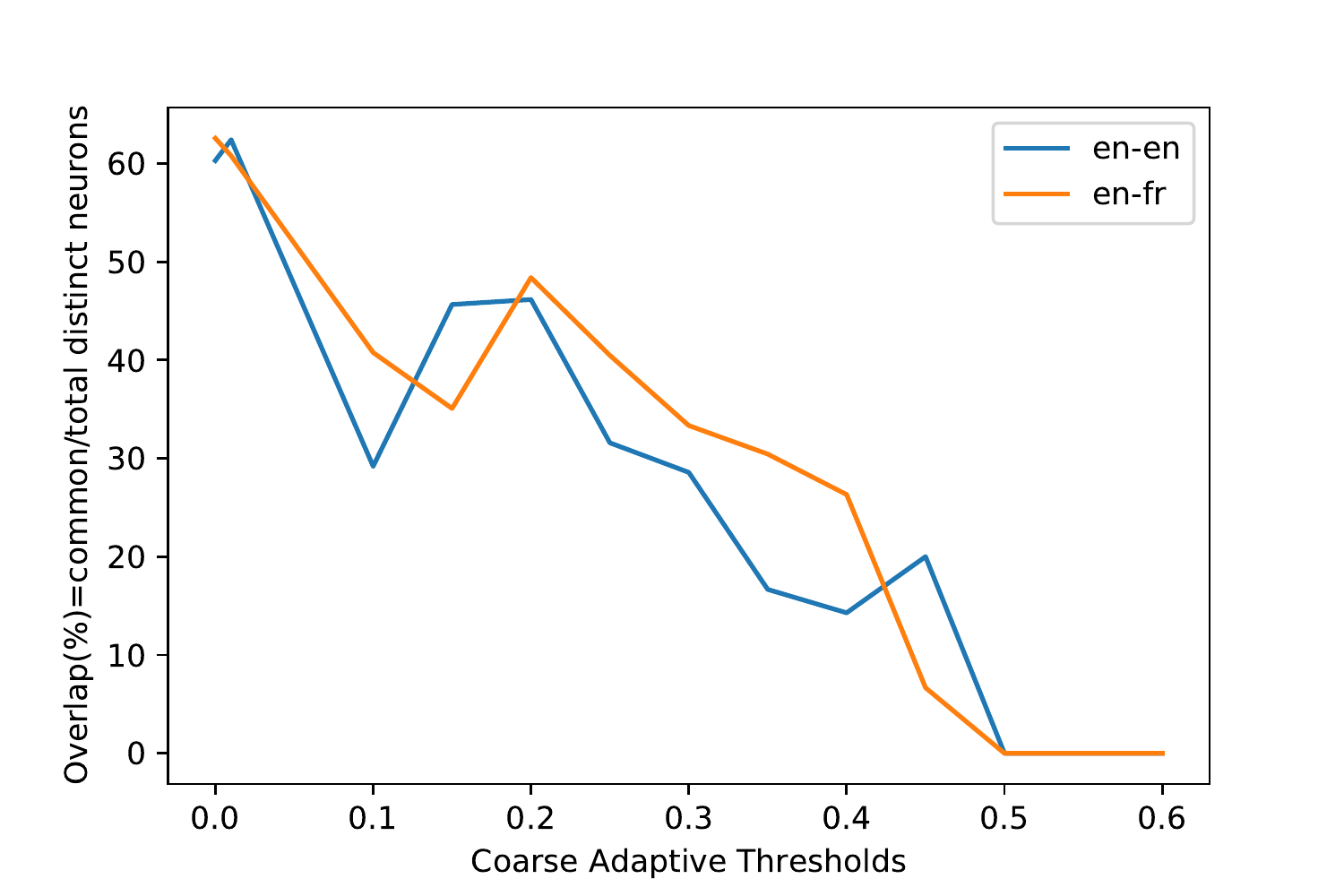}
    \caption{Variation of overlap of attributed neurons for facts in different languages and same languages with varying $t$.}
    \label{fig:diff_lang_same_fact}
\end{figure}

We calculate the overlap between the neurons used for the two sets of English prompts and plot it in \textcolor{cyan}{blue}. We also measure the overlap between the neurons selected for the French prompts and those for the combined set of English prompts and plot this overlap in \textcolor{orange}{orange}, in Figure~\ref{fig:diff_lang_same_fact}. We observe that both overlaps show similar trends and are close to each other. In general, as expected, they show a falling trend as we increase $t$. This closeness of the two curves in Figure~\ref{fig:diff_lang_same_fact} is an attestation to the ability of transformers to efficiently store multilingual information, by taking information in different languages to a common representation space (indicated by the common neurons). This provides further force to the claim of \cite{maksym_21}, that different languages do share a common representation space.

\begin{figure}
    \centering
     \begin{subfigure}[b]{0.3\textwidth}
         \centering
         \includegraphics[width=\textwidth]{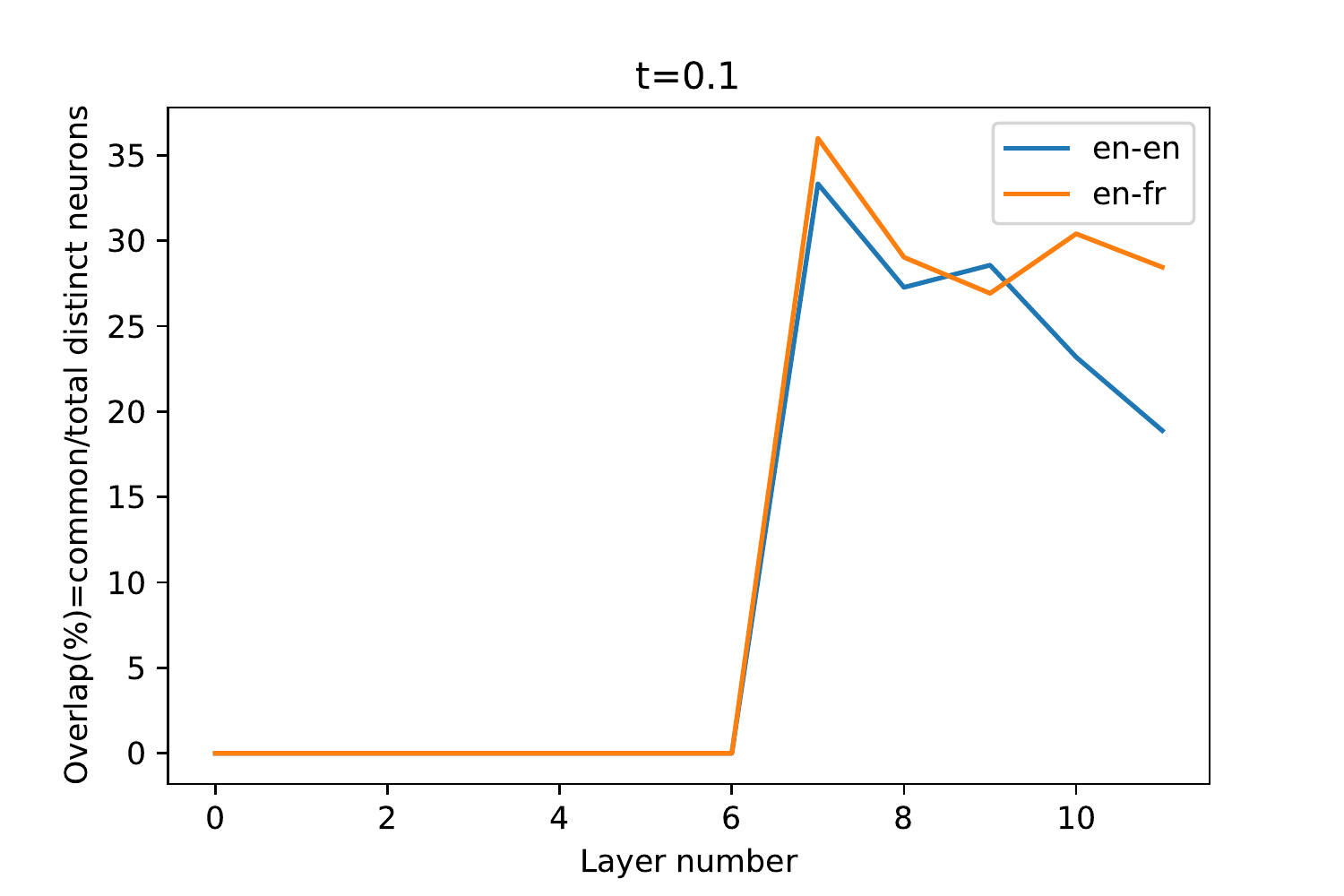}
         \caption{$t=0.1$}
         \label{fig:overlap_1_1}
     \end{subfigure}
     \hfill
     \begin{subfigure}[b]{0.3\textwidth}
         \centering
         \includegraphics[width=\textwidth]{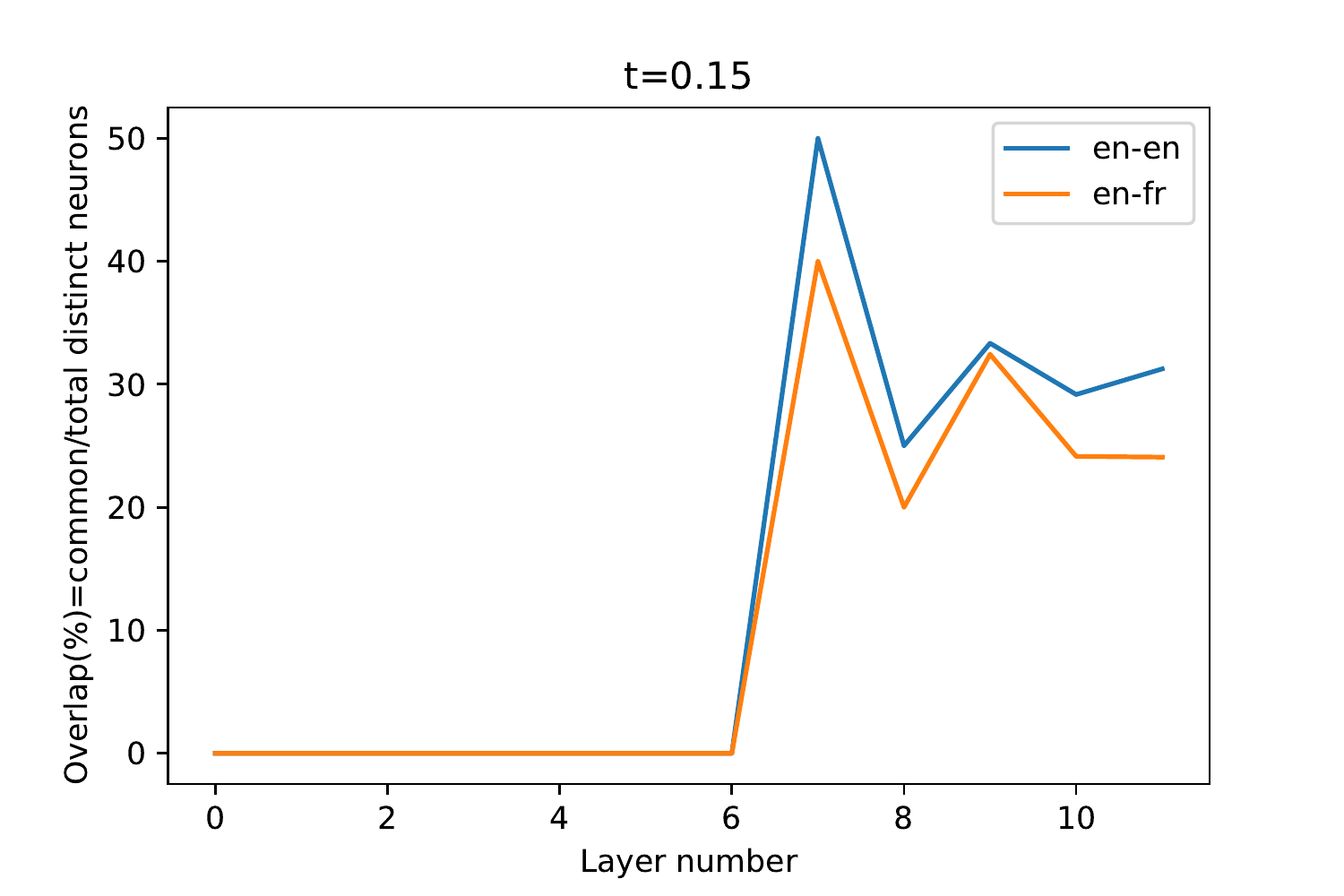}
         \caption{$t=0.15$}
         \label{fig:overlap_2_1}
     \end{subfigure}
     \hfill
     \begin{subfigure}[b]{0.3\textwidth}
         \centering
         \includegraphics[width=\textwidth]{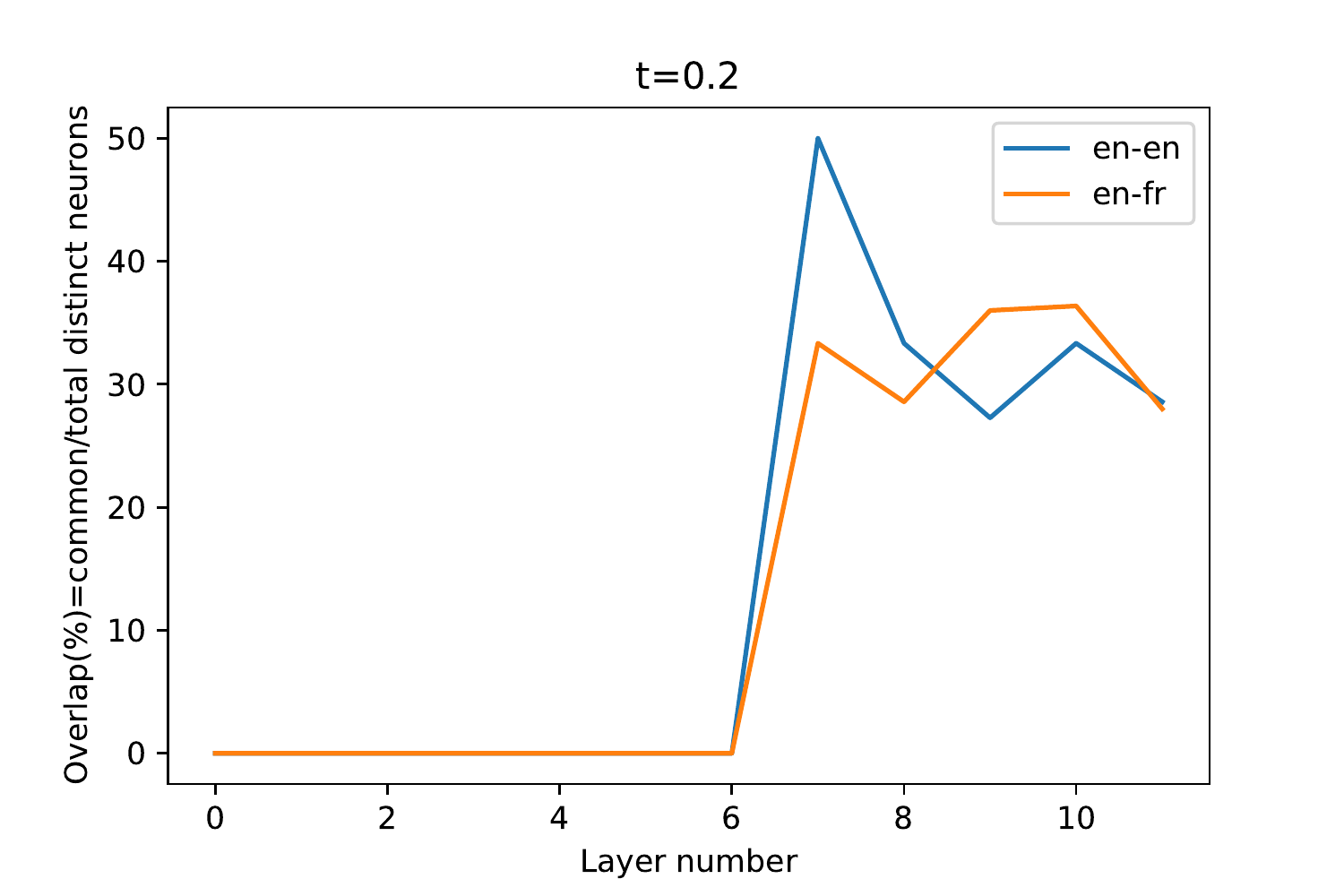}
         \caption{$t=0.2$}
         \label{fig:overlap_3_1}
     \end{subfigure}
     \hfill
     \begin{subfigure}[b]{0.3\textwidth}
         \centering
         \includegraphics[width=\textwidth]{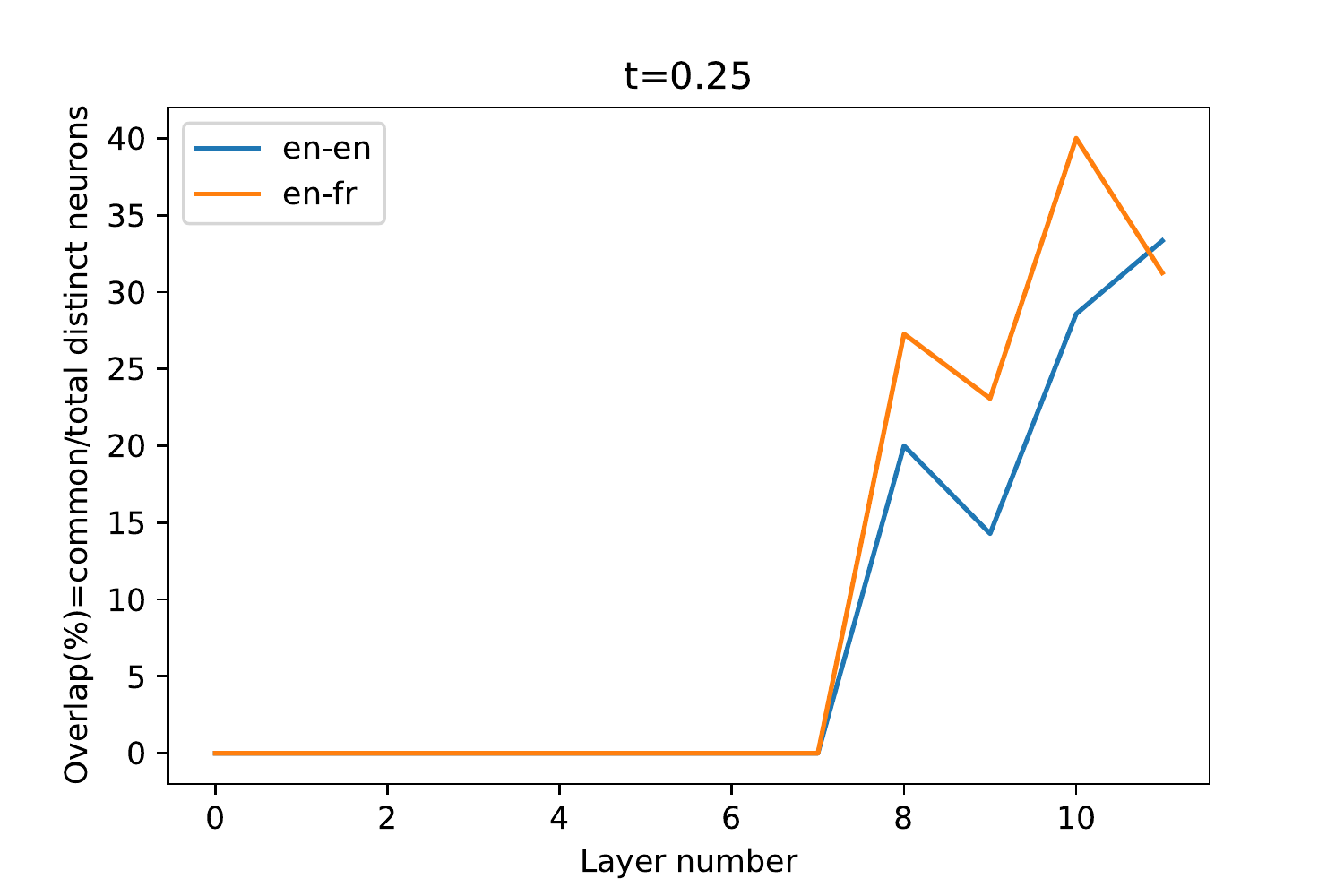}
         \caption{$t=0.25$}
         \label{fig:overlap_4_1}
     \end{subfigure}
     \hfill
     \begin{subfigure}[b]{0.3\textwidth}
         \centering
         \includegraphics[width=\textwidth]{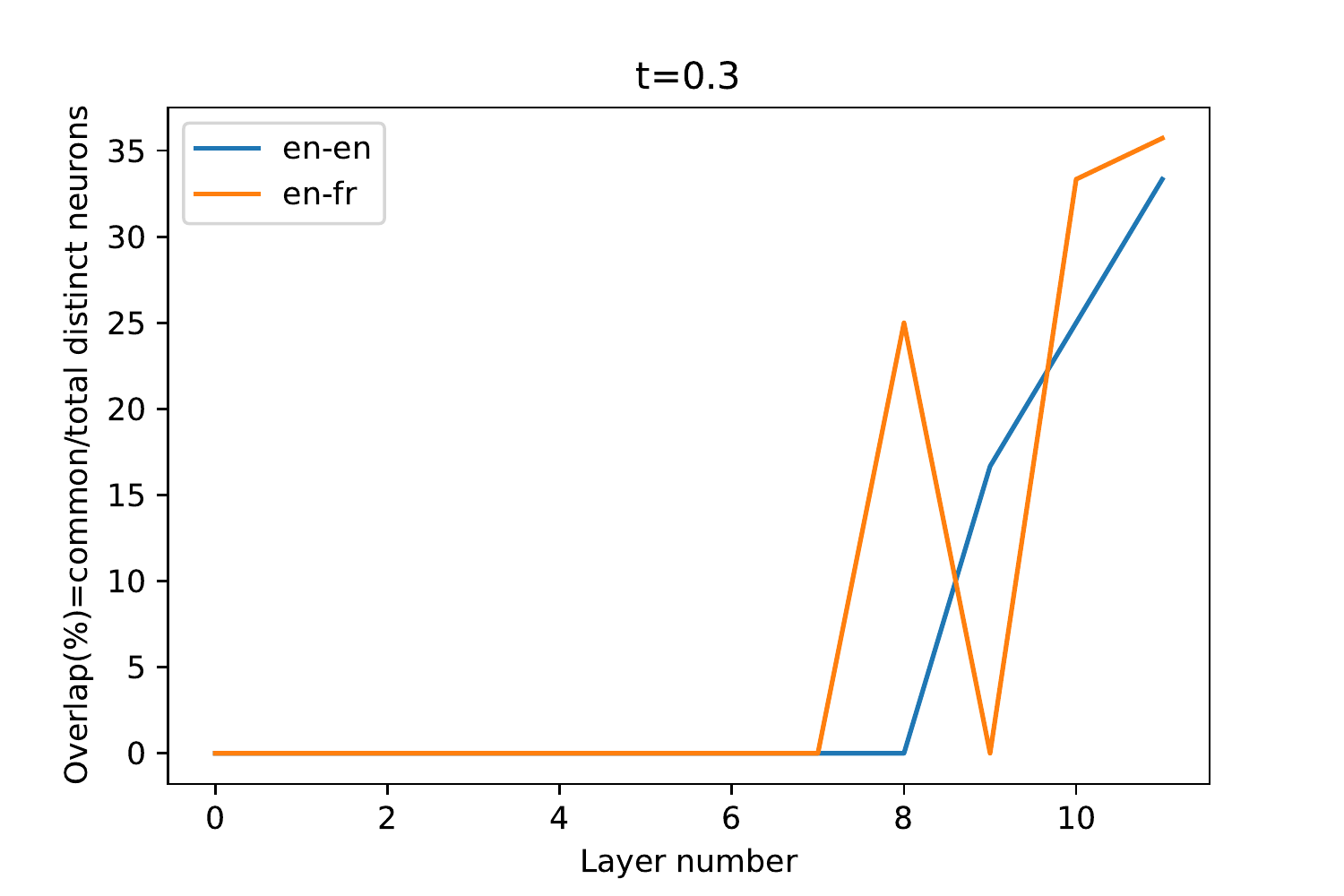}
         \caption{$t=0.3$}
         \label{fig:overlap_5_1}
     \end{subfigure}
    \hfill
     \begin{subfigure}[b]{0.3\textwidth}
         \centering
         \includegraphics[width=\textwidth]{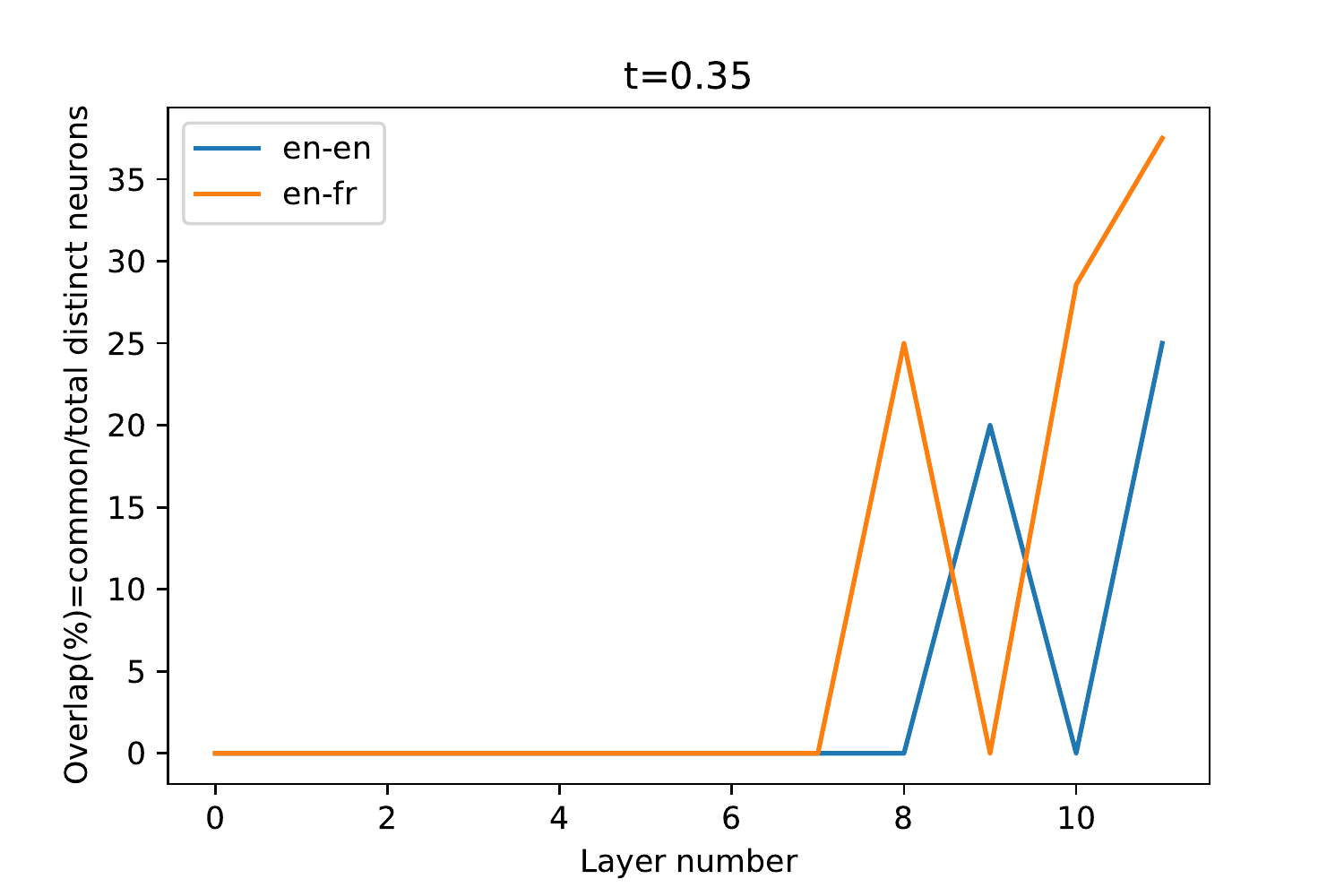}
         \caption{$t=0.35$}
         \label{fig:overlap_6_1}
     \end{subfigure}
     \hfill
     \begin{subfigure}[b]{0.3\textwidth}
         \centering
         \includegraphics[width=\textwidth]{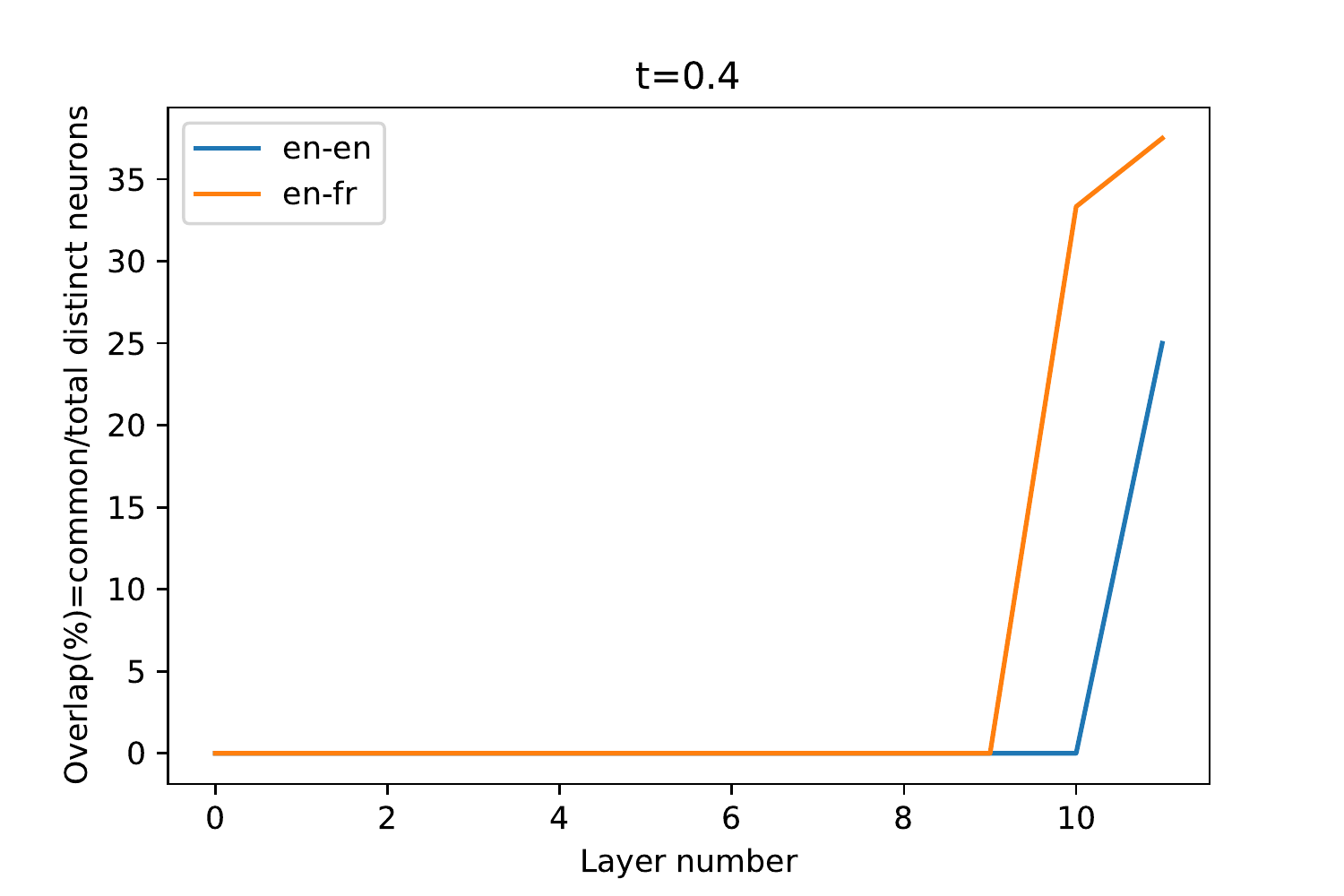}
         \caption{$t=0.4$}
         \label{fig:overlap_7_1}
     \end{subfigure}
     \hfill
     \begin{subfigure}[b]{0.3\textwidth}
         \centering
         \includegraphics[width=\textwidth]{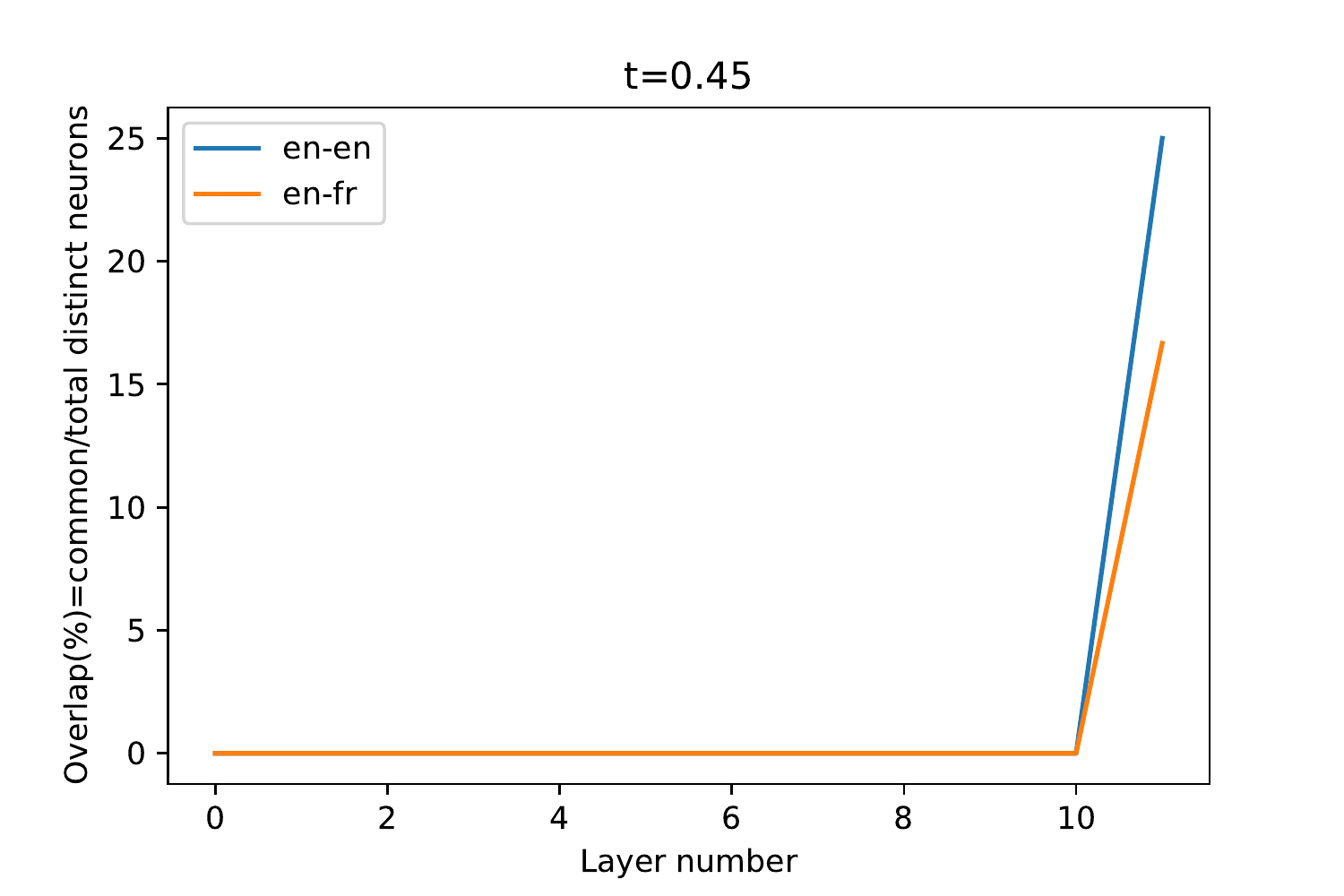}
         \caption{$t=0.45$}
         \label{fig:overlap_8_1}
     \end{subfigure}
     \hfill
     \begin{subfigure}[b]{0.3\textwidth}
         \centering
         \includegraphics[width=\textwidth]{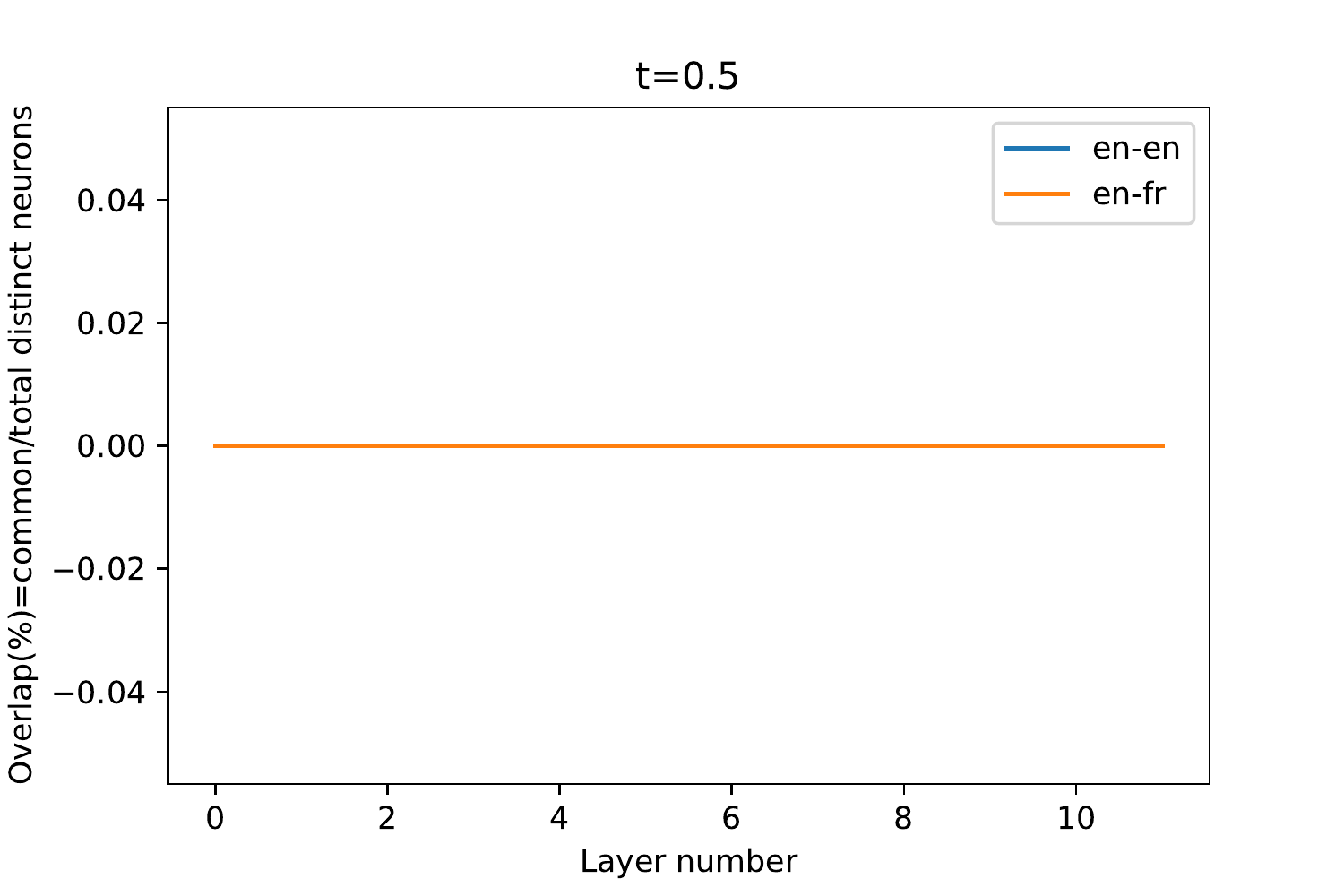}
         \caption{$t=0.50$}
         \label{fig:overlap_9_1}
     \end{subfigure}
      \caption{Layer-wise overlap of neurons between facts in same language and different language at different $t$.}
    \label{fig:diff_lang_overlap_at_layers_with_t}
\end{figure}

The layer-wise analysis is shown in Figure~\ref{fig:diff_lang_overlap_at_layers_with_t}. For $t \in [0.1, 0.25]$ we observe the same trend as before(Figure~\ref{fig:relation_or_fact_fr_de}), where overlap increases first, then decreases in the last few layers. A difference we observe is that the overlap stays non-zero for much higher $t$'s than the ones we obtained while comparing measuring overlap for (France, Capital, Paris) and (Germany, Capital, Berlin). This indicates that the same facts in different languages share many more common neurons of {\bf m-BERT}, than those shared between facts representing the same relation. 

For $t>0.25$, the trend is different and we see no drop in higher layers, probably because we only choose neurons with very high attribution scores, which means that they may already correspond to very specialized information. At around $0.5$, we see that no overlap exists between the neurons used for different languages. 

\section{Grammatical Knowledge Attribution}

As we have prompts checking for factual knowledge attribution, we can also design prompts for grammatical knowledge attribution. We investigate the properties, of grammatical knowledge attribution next. We use the Number Agreement dataset curated by \cite{colorless-green}. We use these data with BERT, in a way similar to what \cite{bert-syntax} do in their repository\footnote{\url{https://github.com/yoavg/bert-syntax}}. 

As this dataset consists mainly of non-sensical sentences and lacks semantic or lexical cues, we can be sure that we are indeed considering attribution of grammatical knowledge. Moreover, as there are correct(\textbf{good}) labels, as well as incorrect(\textbf{bad}) ones(corresponding to the wrong number of subject/verb) provided, we can check which neurons are responsible for the correct prediction, and which push the prediction in the wrong direction. We also show variation of results as the number of attractors(words that distract the model from predicting the correct number) present in the sample prompt, change. 

\begin{figure}
    \centering
    \begin{subfigure}{0.45\textwidth}
        \includegraphics[width=\textwidth]{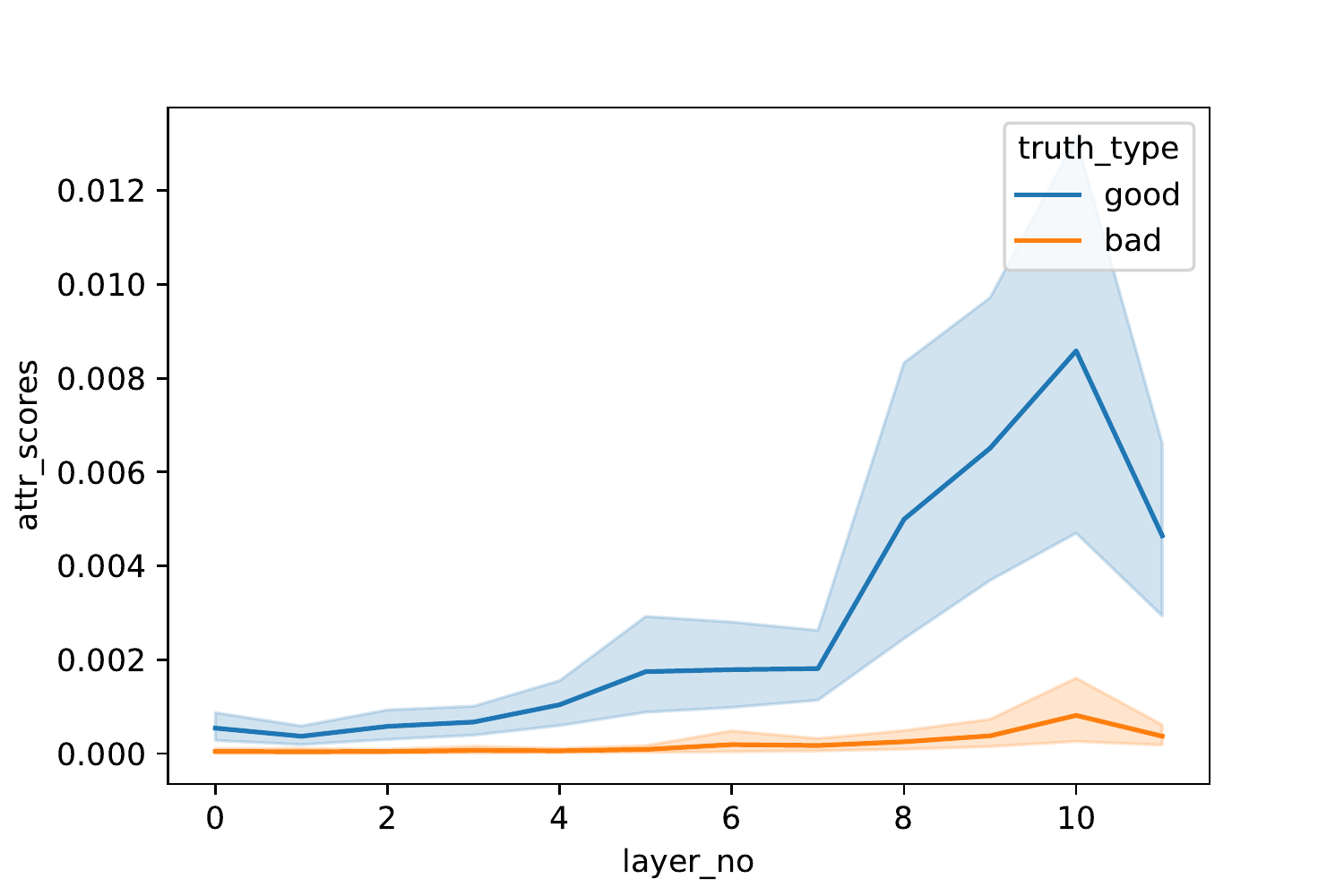}
        \caption{Max. attribution scores in each layer}
        \label{fig:truth_based_gka_subfig1}
    \end{subfigure}
    \hfill
    \begin{subfigure}{0.45\textwidth}
        \includegraphics[width=\textwidth]{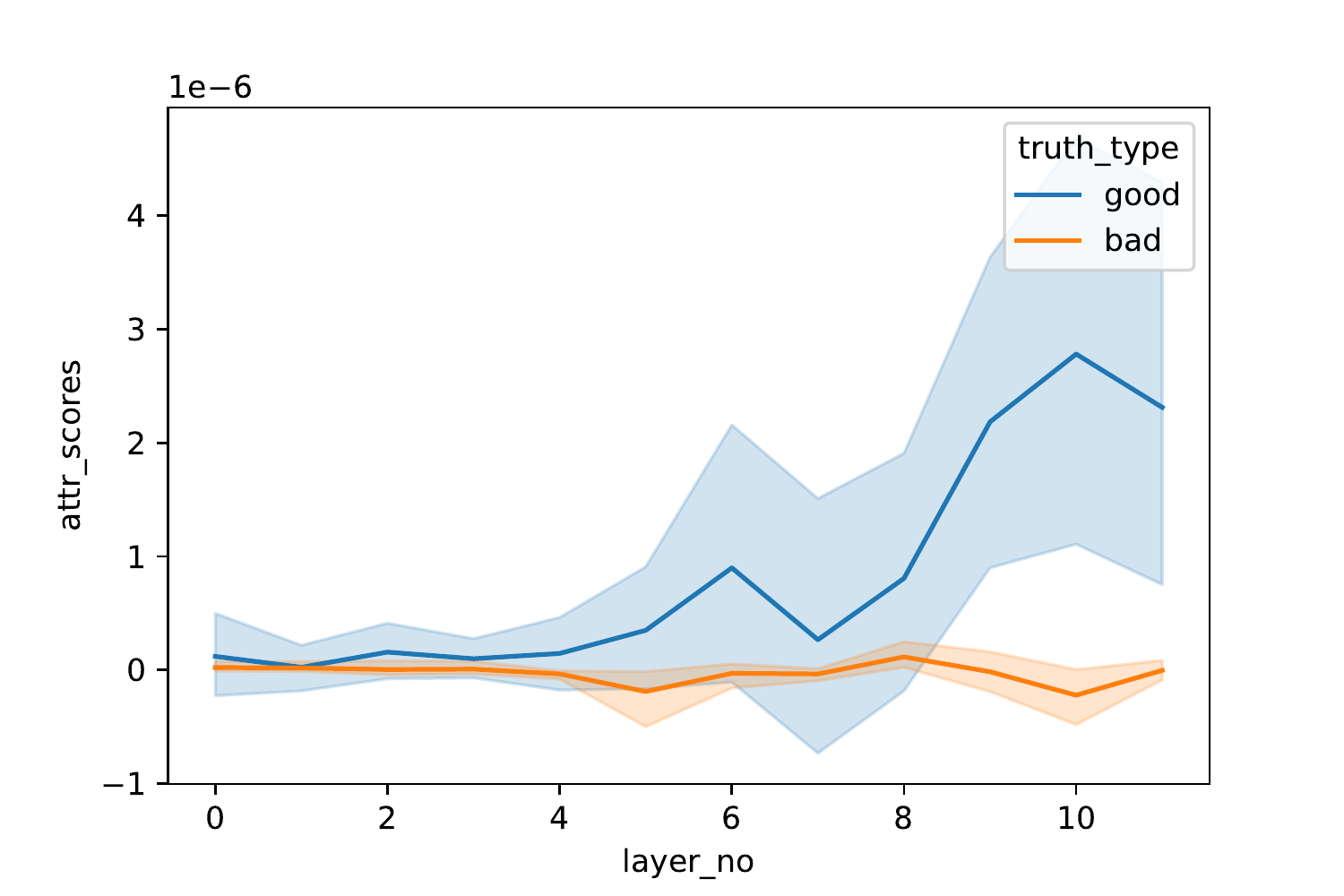}
        \caption{Mean attribution scores in each layer}
        \label{fig:truth_based_gka_subfig2}
    \end{subfigure}
    \caption{
    Variation of attribution scores for predicting correct number(in \textcolor{blue}{blue}) and incorrect number(in \textcolor{orange}{orange}). The scores are averaged over all examples in the dataset and shaded regions indicate confidence intervals.}
    \label{fig:truth_based_gka}
\end{figure}

Figure~\ref{fig:truth_based_gka} shows the layer-wise variation of attribution scores for predicting the word with the correct number, and predicting the word with incorrect number, at the masked position. 

First, we observe that the attribution scores for correct predictions are almost always higher than the attribution scores for incorrect predictions, showing the ability of the model to discriminate between the two, beginning as early as the $4^{th}$ layer. Second, we see that the highest attribution scores in the case of grammatical knowledge are around 0.1, compared to around 0.3 in the case of factual knowledge(Figure~\ref{fig:max_scores_fr} and Figure~\ref{fig:max_scores_de}). This probably is an indication that grammatical knowledge is more dispersed in the model than factual knowledge.

\begin{figure}
    \centering
    \begin{subfigure}{0.45\textwidth}
        \includegraphics[width=\textwidth]{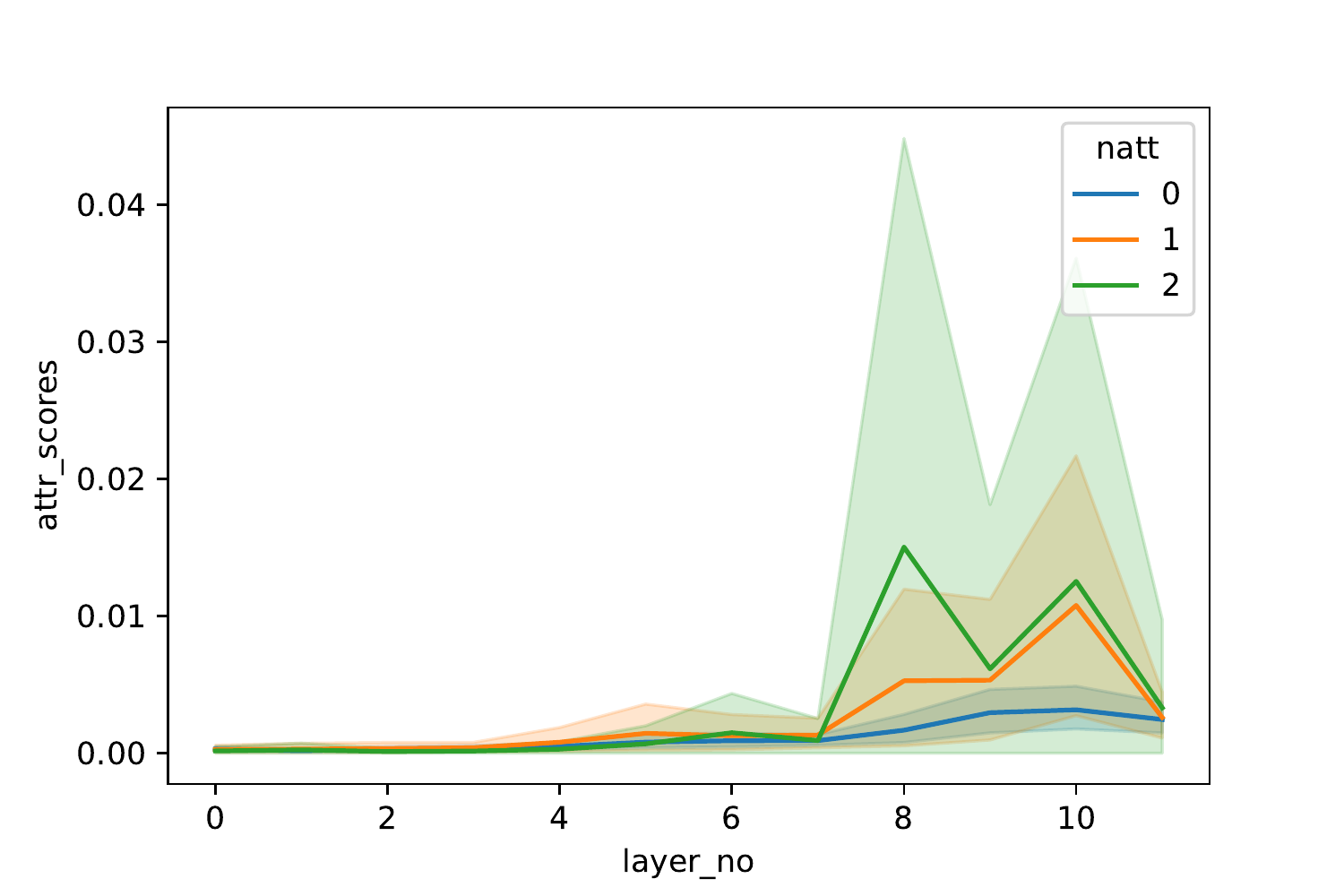}
        \caption{Max. attribution scores in each layer}
        \label{fig:natt_based_gka_subfig1}
    \end{subfigure}
    \hfill
    \begin{subfigure}{0.45\textwidth}
        \includegraphics[width=\textwidth]{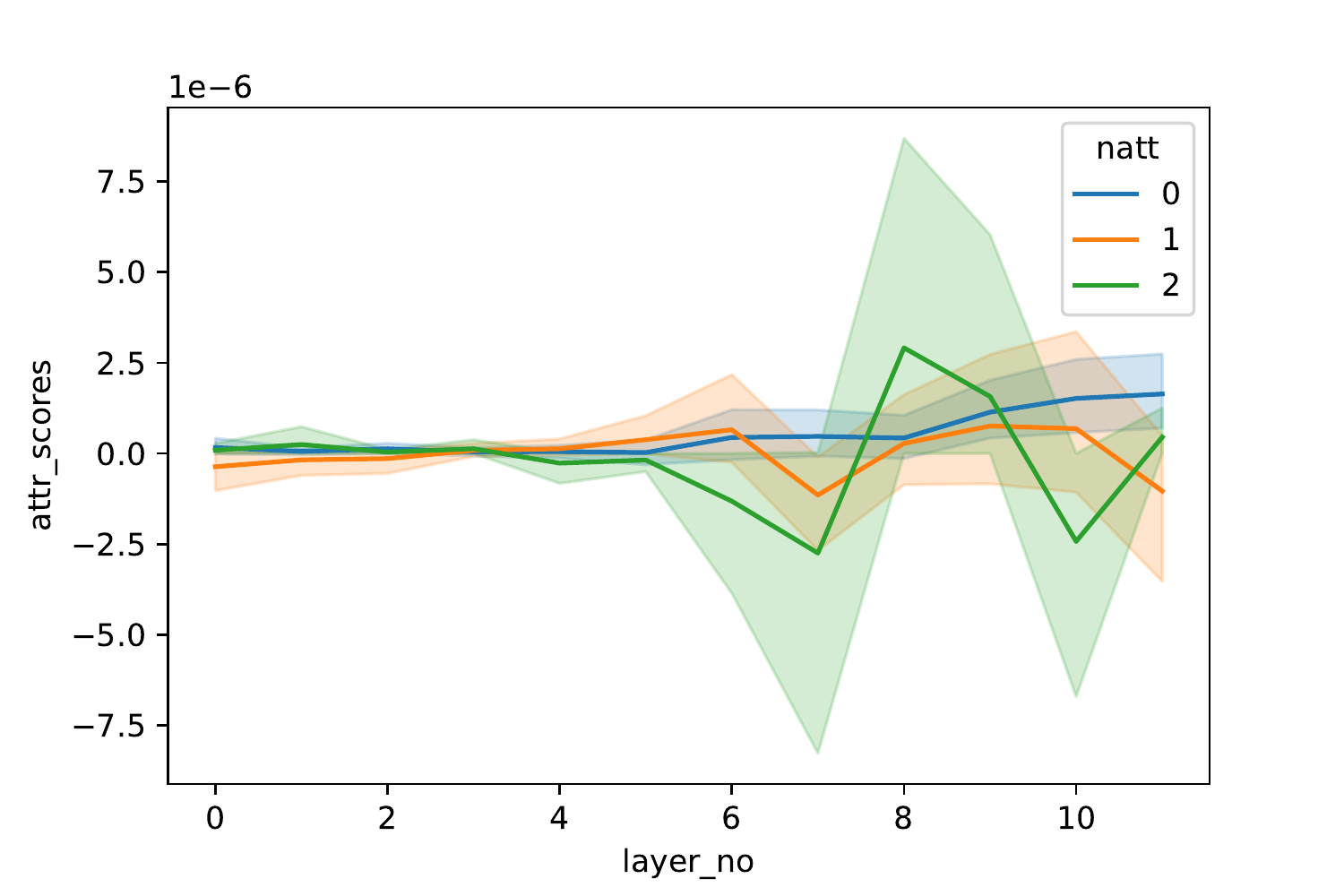}
        \caption{Mean attribution scores in each layer}
        \label{fig:natt_based_gka_subfig2}
    \end{subfigure}
    \caption{
    Variation of attribution scores for examples with different number of attractors. The scores are averaged over all examples in the dataset and shaded regions indicate confidence intervals.}
    \label{fig:natt_based_gka}
\end{figure}

Figure~\ref{fig:natt_based_gka} shows the variation of the attribution scores for examples categorised according to various number of attractors present. We observe that the variation of attribution scores increases significantly as we increase the number of attractors, showing that the model is indeed getting confused by the presence of attractors. In addition, the variation increases as we go to higher layers. 

\begin{figure}
    \centering
    \begin{subfigure}{0.45\textwidth}
        \includegraphics[width=\textwidth]{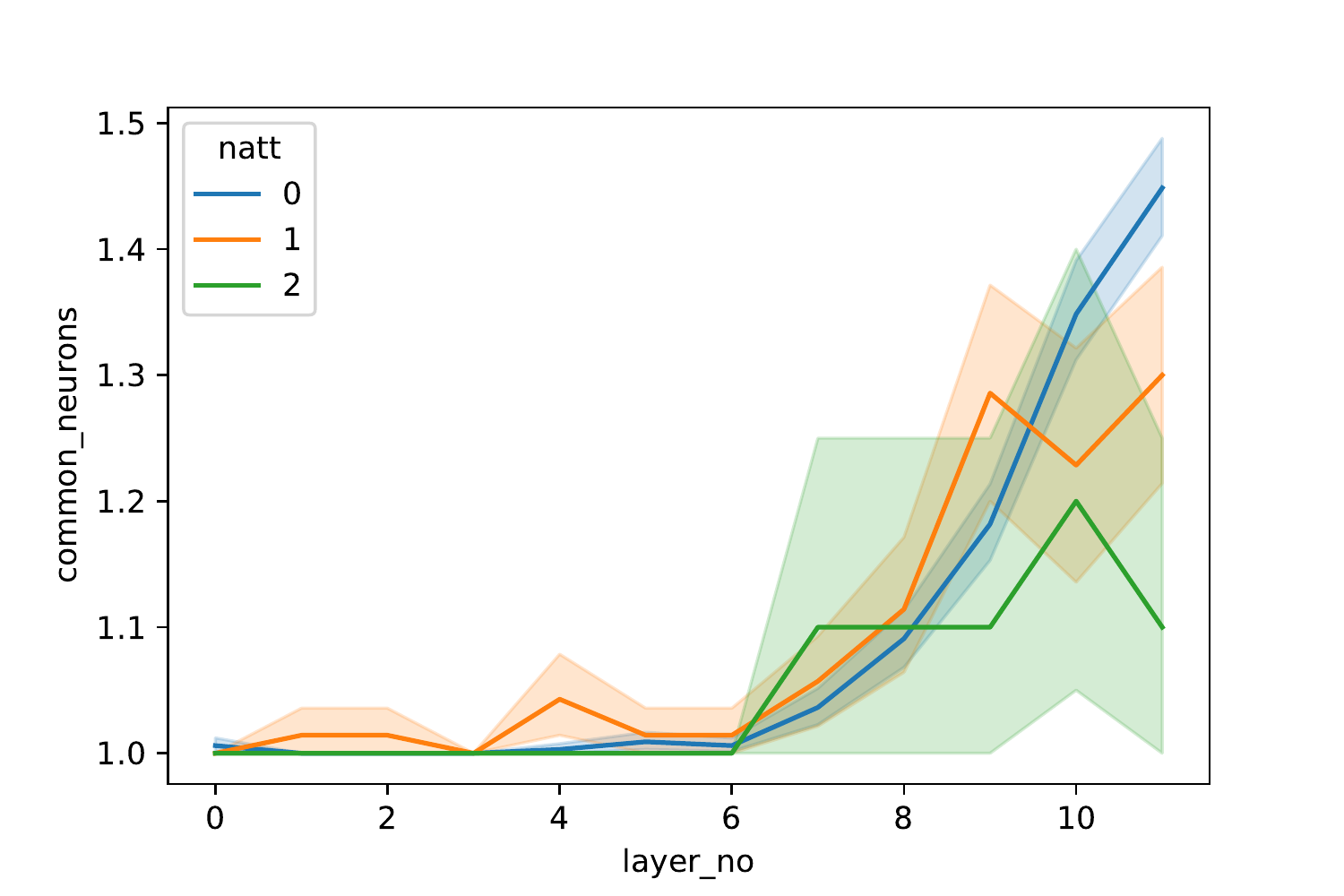}
        \caption{\scriptsize Common refined neurons for correct and incorrect prediction of the same sample}
        \label{fig:cdn_subfig1}
    \end{subfigure}
    \hfill
    \begin{subfigure}{0.45\textwidth}
        \includegraphics[width=\textwidth]{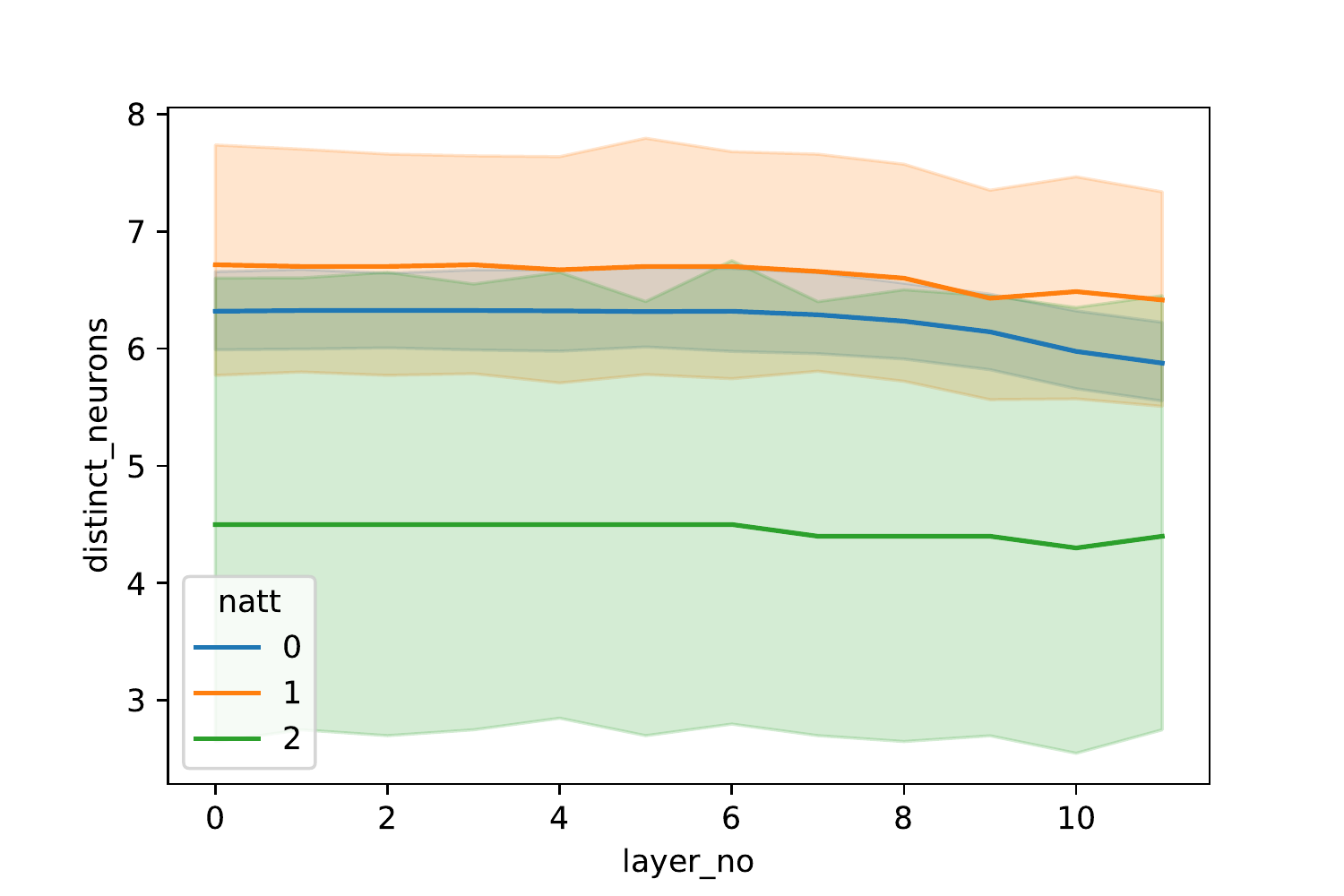}
        \caption{\scriptsize Distinct refined neurons for correct and incorrect prediction of the same sample}
        \label{fig:cdn_subfig2}
    \end{subfigure}
    \caption{
    Variation of number of refined neurons for examples with different number of attractors. The numbers are averaged over all examples in the dataset and shaded regions indicate confidence intervals.}
    \label{fig:common_distinct_neurons}
\end{figure}

Next, we find refined neurons for the correct and incorrect prediction of each sample in the dataset, using an adaptive threshold of 0.5. We calculate the number of neurons common for the correct and incorrect predictions(Figure~\ref{fig:cdn_subfig1}) and number of neurons that store information for either the correct answer or the incorrect answer("decided" neurons). The common neurons("undecided" neurons) indicate those neurons that do not differentiate between the correct and incorrect predictions, while the distinct ones are the ones which only support only one of correct or the incorrect answer. We see that these "undecided" neurons occur more towards the later layers(Figure~\ref{fig:cdn_subfig1}) but are much less in number, than the "decided" neurons shown in Figure~\ref{fig:cdn_subfig2}.

\section{Conclusion and Future Work}
We presented various interesting patterns on the location of knowledge of various kinds in various parts of transformer networks and the interactions of various kinds and languages in which knowledge can occur using the Knowledge Neurons\citep{dai2021knowledge} attribution technique. 

We leave it for future work to explore other attribution techniques(\cite{lime}, \cite{shapley}, \cite{algebraist_2020}) and on other language models like GPT\citep{gpt}, which have been trained with a different objective(CLM) than the MLM objective of BERT model we have used.

\bibliography{citations.bib}


\begin{appendices}
\section{Prompts used}
\label{appendix_first}
\subsection{English Prompts-(France, Capital, Paris)-Set 1}

\begin{enumerate}
    \item "Sarah was visiting [MASK], the capital of france",
    
    \item "The capital of france is [MASK]",
    
    \item "[MASK] is the capital of france",
    
    \item "France's capital [MASK] is a hotspot for romantic vacations",
    
    \item "The eiffel tower is situated in [MASK]",
    
    \item "[MASK] is the most populous city in france",
    
    \item "[MASK], france's capital, is one of the most popular tourist destinations in the world"
\end{enumerate}

\subsection{English Prompts- (France, Capital, Paris)- Set 2}
\begin{enumerate}
    \item "I have always wanted to know what the capital of France is... and now I know it is [MASK] !",
    
    \item "Do you even know the name of the capital of France? It's not Berlin, it's [MASK].",
    
    \item "The city of [MASK] was made the capital of France in 987 A.D.",
    
    \item "Claude Monet, the famous painter was also born in [MASK], the now capital of France.",
    
    \item "As is London the gem of Britain, so is [MASK], the centre of France.",
    
    \item "At the beginning of 20th century, [MASK], being the capital of France, was the largest catholic city in the world.",
    
    \item "The city of [MASK], housing the largest business district in Europe, La Défense, is the rightful of its place as the capital of France.",
    
\end{enumerate}

\subsection{French Prompts-(France, Capitale, Paris)}
\begin{enumerate}
    \item "Sarah était en visite à [MASK], la capitale de la France",
    
    \item  "La capitale de la france est [MASK]",
    
    \item "[MASK] est la capitale de la France",
    
    \item "[MASK], la capitale de la France, est un haut lieu des vacances romantiques",
    
    \item "La tour eiffel est située à [MASK]",
    
    \item "[MASK] est la ville la plus peuplée de France",
    
    \item "La capitale de la France, [MASK], est l'une des destinations touristiques les plus populaires au monde",
\end{enumerate}

\subsection{English Prompts-(Germany, Capital, Berlin)}
\begin{enumerate}
    \item "Why would Sarah ever visit [MASK]? Doesn't she know that its the capital of Germany!?",
    \item "Hitler lived most of its life in [MASK], the now capital of Germany.",
    \item "[MASK] is the capital of Germany",
    \item "Germany's capital [MASK] is a hotspot for tourists",
    \item "The Brandenburg Gate is situated in [MASK]",
    \item "[MASK] is the most populous city in Germany",
    \item "[MASK], Germany's capital, is visited by a lot of Dutch people every year",
\end{enumerate}

\subsection{English Prompts-(Cow, Eats, Grass)}
\begin{enumerate}
    \item "Yesterday, I saw a cow eating [MASK]",
    \item "Cows love to eat [MASK]",
    \item "Herbivores like cows, that eat [MASK], usually have a separate stomach to ruminate",
    \item "All the nutrients in cow's milk comes from the [MASK] it eats",
    \item "Some people were found to be living worse than cows: eating breads made out of [MASK]",
    \item "The cows in praries look most beautiful when gnawing on [MASK], just before the sunset",
    \item "Spherical cows eat get hungry too! They too eat [MASK]."
\end{enumerate}
\end{appendices}

\end{document}